\pdfoutput=1

\PassOptionsToPackage{dvipsnames}{xcolor}

\documentclass{article} % For LaTeX2e
\usepackage[utf8]{inputenc}
\usepackage[T1]{fontenc}   

\usepackage[normalem]{ulem}
\usepackage{booktabs}
\usepackage{pifont}
\usepackage[numbers]{natbib}
\usepackage{xcolor}
\usepackage{algorithmic}
\usepackage{algorithm}
\usepackage{caption}
\usepackage{subcaption}
\usepackage{dsfont}

\newlength{\defbaselineskip}
\usepackage{hyperref}
\usepackage{url}
\usepackage{graphicx}
\usepackage{thm-restate}
\usepackage{multirow}
\usepackage{lineno}
\usepackage{mathtools}
\usepackage[export]{adjustbox}
\usepackage{amssymb}
\usepackage{authblk}
\usepackage{bm}

\usepackage{enumitem}

\allowdisplaybreaks

\newcommand{\michael}[1]{}
\newcommand{\michaeladdressed}[1]{}
\newcommand{\malcolm}[1]{}

\begin{document}
\title{
Using Pre-trained LLMs for Multivariate Time Series Forecasting
%Targeted Fine-Tuning of LLMs for \\ Multivariate Time Series Forecasting
%Fine-Tuning LLMs for Multivariate Time Series Forecasting
%\\ OR \\
%Parameter-Efficient Fine-Tuning of LLMs to Learn Embeddings for Multivariate Time Series Forecasting
%\\ OR \\
%Layer-Normalization Fine-Tuning of LLMs to Learn Embeddings for Multivariate Time Series Forecasting
%\\ OR \\
%Partial Fine-Tuning of LLMs to Learn Embeddings for Multivariate Time Series Forecasting
%\\ OR \\
%Using LLMs to Learn Embeddings for Multivariate Time Series Forecasting
%Fine-Tuning LLMs for Time Series Forecasting
}

\author[1]{Malcolm L. Wolff}
\author[2]{Shenghao Yang}
\author[1]{Kari Torkkola}
\author[1]{Michael W. Mahoney}
\affil[1]{Amazon Supply Chain Optimization Technologies (New York, NY 10001)}
\affil[2]{University of Waterloo}
\affil[1]{ {\texttt{\{wolfmalc, karito, zmahmich}\}\texttt{@amazon.com}} }
\affil[2]{ {\texttt{s286yang@uwaterloo.ca}} }

\date{} %\date{DRAFT 01/02/24 --- DO NOT DISTRIBUTE}

\maketitle

\begin{abstract}
\noindent
Pre-trained Large Language Models (LLMs) encapsulate large amounts of knowledge and take enormous amounts of compute to train.  
We make use of this resource, together with the observation that LLMs are able to transfer knowledge and performance from one domain or even modality to another seemingly-unrelated area, to help with multivariate demand time series forecasting. 
%In particular, we apply LLMs that have been pre-trained for language generation to univariate as well as %multivariate demand time series forecasting problems. 
Attention in transformer-based methods requires something worth attending to -- more than just samples of a time-series. 
We explore different methods to map multivariate input time series into the LLM token embedding space. 
In particular, our novel multivariate patching strategy to embed time series features into decoder-only pre-trained Transformers produces results competitive with state-of-the-art time series forecasting models.
We also use recently-developed weight-based diagnostics to validate our findings.
\end{abstract}

%\newpage
%\vspace{-3mm}
\section{Introduction}
%\vspace{-2mm}

Time series forecasting refers to a class of techniques for the prediction of events through a sequence of time, typically to inform strategic or tactical decision making.
Going beyond strategic forecasting problems (e.g., those commonly-used historically in statistics and econometrics~\citep{harvey90}), operational forecasting problems are increasingly-important. 
%%throughout Amazon and beyond, 
For example, 
at large internet retail companies, this includes demand forecasting for products at an online retailer, work force cohorts of a company in its locations, compute capacity needs per region and server type, etc.;
in scientific machine learning, this includes prediction of extreme events in, e.g., climate and weather models; 
and so on.
In particular, MQCNN~\citep{wen2018multihorizon} and MQTransformer~\citep{eisenach2020mqtransformer} are state-of-the-art (SOTA) neural network (NN) based multivariate time series forecasting models that are used 
%%by SCOT Forecasting 
to predict future demand at the product level for hundreds of millions of products.

Along a seemingly-different direction, Large Language Models (LLMs) exhibit 
multi-modal capabilities, 
broadening the horizon of their potential applicability~\citep{gong2023multimodal, gao2023llama}. 
LLMs appear to exhibit ``emergent behavior'' as they scale, e.g., in the sense that they may exhibit a capacity to execute tasks that are seemingly quite different than the tasks on which they were directly trained~\citep{hahn2023theory, hagendorff2023machine}. 
Motivated by this, and by their promise to serve as a foundation for model development more generally~\citep{foundationStanford_TR}, 
researchers have found that LLMs improve a number of seemingly-different tasks, 
including vision-language tasks~\citep{sung2022vl, chen2022simple, zhu2023minigpt}, 
chain-of-thought reasoning~\citep{zeng2022socratic, zhang2023multimodal}, and instruction tuning~\citep{liu2023visual}.

\paragraph{Our Contribution.}

We evaluate the efficacy pre-trained LLMs for multi-horizon forecasting with multidimensional time-series inputs. 
In particular, we describe a method for the targeted fine-tuning of a small proportion of parameters in an LLM---namely, their layer norms---for use in forecasting multivariate and multi-horizon time series data, and we evaluate this method in comparison with a SOTA time series forecasting baseline.
For our baseline, we use the MQCNN~\citep{wen2018multihorizon} model.
This model
%%, developed at Amazon, 
is a convolutional Seq2Seq architecture, and it massively improved accuracy for retail product demand forecasting, marking the move to NN-based learning for such models.
%and related SOTA multivariate time series forecasting problems.
This model has remained SOTA for retail demand forecasting until very recently, where improvements were made by introducing encoder-decoder attention and decoder self-attention~\citep{eisenach2020mqtransformer}, improving temporal context-alignment and reducing excess variation of the forecast. 
Full Transformer stacks have yet to unanimously overtake simpler encoding mechanisms in the field of time series forecasting \citep{zeng2023transformers}.
Yet, the abundance of recent  Transformer-based methodologies for time series forecasting \citep{kitaev2020reformer, zhou2021informer, wu2022autoformer, zhou2022fedformer} suggests a persistent belief in the usefulness of self-attention mechanisms for time series analysis, given an appropriate embedding space.

\begin{figure}
    \centering
    \includegraphics[scale=.55]{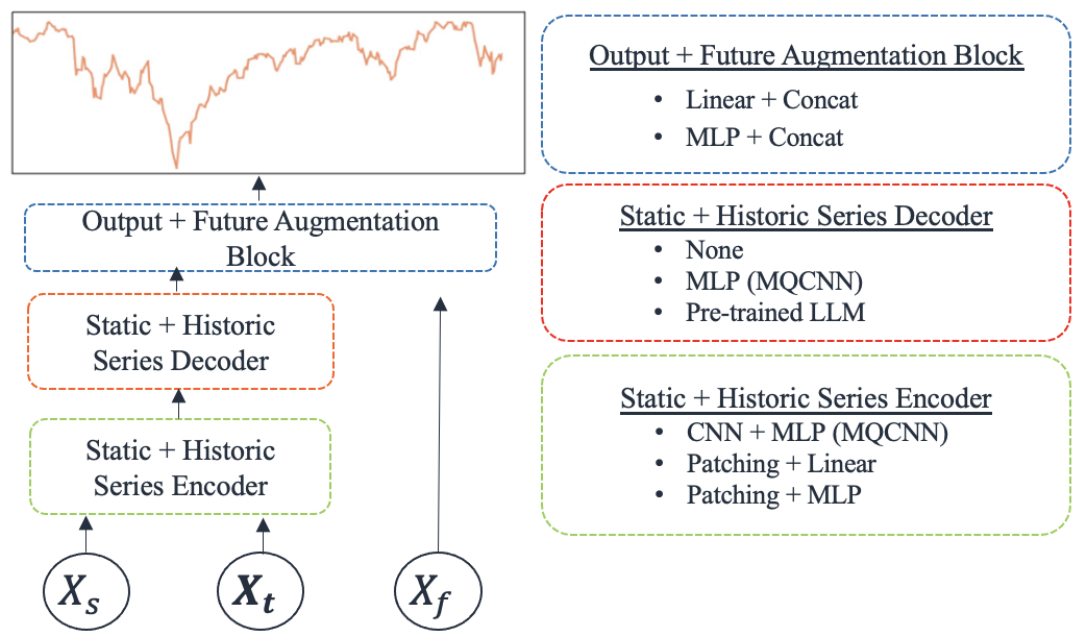}
    \caption{High level architectural design of our experiments. 
    The static + historic series decoder block MLP (MQCNN) is in fact a complex collection of MLPs for different forecasting horizons and quantiles. 
    Pre-trained LLMs considered are GPT-2, Flan-T5, and MPT-7B.
             }
    \label{fig:arch_tsfpt}
\end{figure}

In this paper, we introduce and evaluate methods to map \textit{multivariate input time series} into the token space of pre-trained LLMs, thereby using pre-trained LLMs for multivariate time series forecasting.
This follows and extends recent work in univariate time-series forecasting with the aid of LLMs~\citep{zhou2023fits,chronos_TR}. 
Our approach involves learning simple linear or two-layer MLP embedding maps from the
time series space into the token embedding space;
and, after the LLM, the reverse map from the token embedding space back to time-series space. 
See Figure~\ref{fig:arch_tsfpt} for an illustration.
% COMMENT
% The figure does not  illustrate what is said above (like the Chronos paper figure does). 
% END OF COMMENT
By keeping the LLM weights fixed aside from the layer norms, we drastically reduce the number of trainable parameters, and hence the training time, of the forecasting model.
We also propose \textit{multivariate patching}, extending prior work on LLM fine-tuning \citep{zhou2023fits} to multivariate inputs and multivariate outputs.
We evaluate our model across multiple specifications and pre-trained LLMs, finding our results remain consistent.
We provide an empirical evaluation of these two approaches for LLM-based time series forecasting, using product demand data from a large internet retailer.
\michael{Clarify which ``two approaches'' (linear and two layer?).}
Our comparison baseline is a variant of an existing production forecasting system.
Among other things, we show that fine-tuning a small number of parameters in publicly-available pre-trained LLMs (i.e., just their layer norms) can reach nearly comparable performance to highly-specialized architectures for demand forecasting.
Finally, we use layer-specific weight analysis techniques, based on Heavy-Tailed Self-Regularization (HTSR) Theory~\citep{martin2019traditional,martin2020heavy,martin2021implicit,martin2021predicting,yang2022evaluating}, as model diagnostics to analyze our models.
HTSR Theory is based on the idea that well-trained NN models have HT structure in their spectral (eigenvalue) distributions. 
Among other things, we show a relationship between the spectral distribution of the layer weight Gram Matrix and the quality of the time series-to-LLM embedding, in terms of forecast test accuracy.

%\newpage
%\vspace{-3mm}
\section{Background and Related Work}
\label{sxn:background}
%\vspace{-2mm}

\paragraph{Transformers and LLMs.}
SOTA LLM architectures are primarily based on the Transformer architecture~\citep{vaswani2017attention}. 
The original Transformer model consisted of an encoder and decoder, each of which contain multiple layers of self-attention mechanisms and feed forward networks.
For language modeling tasks, however, often only the decoder or encoder is used. 
Open-source LLMs based on the Transformer architecture include BERT~\citep{devlin2019bert}, the GPT models~\citep{radford2019language,brown2020language}, T5~\citep{raffel2020exploring}, Flan-T5~\citep{chung2022scaling}, and MPT \citep{MosaicML2023Introducing}.

\paragraph{Transformers and Time Series.}
While Transformers have been effective for extracting semantic correlations among elements of a long sequence, their efficacy in encoding temporal correlation has been mixed~\citep{wen2022transformers, zeng2023transformers}. 
Recent techniques for incorporating Transformers into time series forecasting have relied on explicit transformations of the time series inputs~\citep{wu2021autoformer, nie2023time}. 
For example,~\cite{wu2021autoformer} rely on frequency representations to extract stronger temporal correlation, and~\cite{nie2023time} segment the time series into aggregated overlapping subseries, or ``patches,'' to explicitly induce local information retention.

\paragraph{LLMs and Time Series.}
In spite of these challenges, recent work has employed transformer-based LLMs for univariate time series forecasting, with surprising success~\citep{zhou2023fits, yu2023temporal, nie2023time}. 
In particular,~\cite{zhou2023fits} explores the efficacy of pre-trained LLMs for time series forecasting by learning linear maps from ``patched'' time series to the input and output of frozen LLMs.
This allows layer norms to be fine-tuned, achieving comparable or improved performance over time series transformer models specialized for time series \citep{kitaev2020reformer, zhou2021informer, wu2022autoformer, zhou2022fedformer}. 
The authors also provide evidence that self-attention performs similarly to principal component analysis, providing intuition for the generalizability of LLMs \citep{zhou2023fits}. 
This approach is named FPT, or Frozen Pretrained Transformer~\citep{zhou2023fits}.

On the other hand, using Flan-T5 as a base LLM model, the Chronos method~\cite{chronos_TR} quantizes a time-series values to 4,096 discrete levels, each represented by a token, and it then fine-tunes all LLM weights. While quite expensive, results showed that in-domain Chronos scores rank better than dataset-specific baselines on all of the 15 benchmark datasets. 
In zero-shot tests on 27 datasets, their model came close to the best dataset specific baseline scores across the board despite not having been trained on these datasets at all. 
A very recent version, Chronos-Bolt, foregoes quantization and instead does univariate patching followed by embedding and a T5 model, the weights of which are fully trained. 
Even a tiny T5 model reaches top performance, which demonstrates the superiority of patching and embedding compared to quantization \cite{fatir2024ChronosBolt}. 

%
% COMMENT: The following was the initial observation but it was later found that random initialization was 
% done incorrectly - when done correctly, there was no difference!
%
%The FlanTS work also shows that, compared to the pretrained LLM as the initialization, random %initialization of the same architecture never converges to low error, indicating the value of language %prediction pre-training for this forecasting task~\citep{fatir2023flants}. 

\paragraph{Weight Diagnostics for Model Analysis.}
The diagnostic tools we use depend on the weight matrices of the trained models.%
\footnote{In particular, since they depend on HTSR Theory~\citep{martin2021implicit}, our metrics do not require access to training/testing data, and they do not need knowledge of training protocols~\citep{martin2021predicting,yang2022evaluating}. This is important when working with~LLMs.}
Weight analysis of NN models has been considered recently~\citep{NIPS2017_a96b65a7, frankle2020early, damian2022neural}.
Most prominent, and most relevant for our approach, is work on so-called Heavy-Tailed Self-Regularization (HTSR) Theory~\citep{martin2019traditional,martin2020heavy,martin2021implicit,martin2021predicting,yang2022evaluating,ww_timeseries_TR}.
HTSR theory 
%%(see two recent AMLC papers~\citep{wolfe-mahoney-amlc22,wolfe-mahoney-amlc23} for details) 
uses ideas from the statistical physics of learning and heavy-tailed (HT) random matrix theory to formalize the idea that well-trained models should have layer-wise weight matrices that have eigenvalue correlations that exhibit some sort of HT structure.
This HT structure can often be modeled by a power law (PL) or truncated power law (TPL) functional form.
Based on this, one can consider the empirical spectral distribution (ESD) of individual layer weight matrices, fitting these distributions to PL and TPL distributions.
From this, one obtains either layer-wise metrics or (by averaging across layers in a model) aggregate metrics for a given model.
These metrics (e.g., $\alpha$, $\lambda_{max}$, and $\hat{\alpha}$ below), quantify the HT shape of the ESD, and they can be used to measure the HT structure in the correlations of a given layer or of an entire model.
%%External publications have shown that t
These metrics can be used to predict the trends in the quality of SOTA Computer Vision (CV) and Natural Language Processing (NLP) models, even without access to training/testing data~\citep{martin2021predicting,yang2022evaluating}; and 
more recent 
%%Amazon-internal 
publications have shown that these metrics can be used to diagnose and predict the quality of MQCNN and MQTransformer models~\citep{ww_timeseries_TR}
%%~\citep{wolfe-mahoney-amlc22,wolfe-mahoney-amlc23}, 
i.e., SOTA NN forecasting models in use 
for demand forecasting. 
%%at Amazon.

%\newpage
%\vspace{-3mm}
\section{Main method}
%\vspace{-2mm}

In this section, we describe our main method for fine-tuning LLMs for time series forecasting.

\subsection{General Forecasting Problem}
%\vspace{-1mm}

Here, we describe the general forecasting problem we consider. 
We denote tensors in boldface, matrices in upper case, and vectors in lower case notation. 
Let $Y \in\mathbb{R}^{N\times T}$ denote $N$ time series of length $T$, $\mathbf{X}^{(t)} \in\mathbb{R}^{N\times T \times d}$ a set of $d$ additional time series features, and ${X}^{(s)} \in \mathbb{R}^{N\times m}$ a set of $m$ static covariates. 
Given a \emph{context length} $C \geq 0$, i.e., the number of past observations used for modeling from the forecast time $t$, and a collection of \emph{horizons} $\mathcal{H}$ to forecast in the future, we wish to generate the conditional forecast, given $\mathbf{X}_{t-C:t}^{(t)} = (X_{t-C}^{(t)},..., X_{t}^{(t)})$, ${Y}_{t-C:t}$, and ${X}^{(s)}$, via the model
\begin{equation}
\begin{aligned}
\hat{{Y}}_{t, \mathcal{H}} = f(Y_{t-C:t}, \mathbf{X}_{t-C:t}^{(t)}, {X}^{(s)} ; \bm{\theta}) ,
\end{aligned}
\label{eq:prediction_problem}
\end{equation}
where $\bm{\theta}$ represents a collection of trainable parameters. 
The parameters are tuned to optimize a loss during training as
$$
\text{Loss}(\bm{\theta}) = \sum_i \sum_h \sum_t \ell(y_{i, t, h}, \hat{y}_{i, t,h}),
$$
where $i$ index individual products, and $\hat{y}_{i,t,h}$ are forecasts of scalar $y_{i,t,j}$ at horizon $h$ for product $i$ at forecast creation date $t$.
In particular, we look at the quantile loss $\ell(y,\hat{y}) = (\tau-1_{y < \hat{y}})(y - \hat{y})$, for quantiles $\tau=.5$ and $\tau=.9$.
%\vspace{-1mm}
\subsection{Modern Forecasting Methods}
%\vspace{-1mm}

\michael{Should this subsection be the final or penaultimate ``paragraph'' in Section~\ref{sxn:background}.}

One SOTA Seq2seq architecture used 
for %%in Amazon 
time series forecasting is based on the MQCNN model~\cite{wen2018multihorizon}, which uses an encoder and decoder to model relationship between its input and output sequences. 
The encoder consists of a multi-layer causal convolutional encoder (similar to WaveNet~\cite{oord2016wavenet}) for historic time series features, and a linear encoder for static features. 
The decoder consists of an MLP on embeddings learned by the encoder, alongside future time series information. 
Notably, while the MQCNN decoder has since been improved with the use of Transformers~\cite{eisenach2020mqtransformer}, the convolutional encoder remains unparalleled for forecast accuracy, at least in the demand forecasting domain, as measured by quantile loss. 

Seq2seq architectures have pervaded the SOTA for forecasting models with temporal dependence structures, including those used in language, vision, and time series prediction tasks. 
Moreover, Transformer blocks~\cite{vaswani2017attention} and the self-attention mechanisms therein have become ``foundational'' techniques in these Seq2seq models to model temporal relationships, achieving superior performance across data modalities~\cite{wen2022transformers}.
As evidence continues to mount for scaled-up LLMs' emergent performance on language-adjacent and non-language tasks~\cite{wei2022emergent, chowdhery2022palm}, we explore whether latent correlations learned by the self-attention layers of LLMs may be informative in time series prediction.
The relative success of transformers in ``foundation'' modeling for natural language \citep{openai2023gpt4} and computer vision \citep{oquab2023dinov2} has spurred significant research in a ``foundation'' time series model. 
However, a persistent difficulty in this subfield is a notable lack of open source time series data. 
There have been a number of recent attempts to circumvent this \cite{zhou2023fits, chang2024llm4ts, jin2024timellm, chronos_TR}, perhaps most notably the recently-developed Chronos model~\cite{chronos_TR}.

One such paper \cite{zhou2023fits} aims to construct a ``foundation'' model for \emph{univariate} time series by learning a linear map between the time series and the token space of a pre-trained transformer stack, where the transformer weights are trained on natural language. 
The authors find that their fine-tuned model is able to achieve near-SOTA performance across a number of publicly-available time series forecasting datasets used for benchmarking. 
However, these results were obtained for \emph{univariate} time series, and thus there is a significant gap between this work and the pragmatic use of fine-tuning LLMs for complex forecasting problems, 
whether those problems come from internet retailer demand forecasting or scientific machine learning forecasting.
%%such as those in Amazon ASIN forecasting. 
While a linear map works well for their univariate time series settings, it's not clear whether such a simple embedding would be appropriate for capturing correlations across both time and co-variates. 
Below we describe our architecture for \textit{multivariate} time series forecasting with LLM fine-tuning.

%\vspace{-1mm}
\subsection{Our Method}
%\vspace{-1mm}
Pre-trained LLMs can be viewed as large autoregressive models, predicting a text token from a history of previous tokens. 
A stream of text is ``tokenized'' into a stream of tokens that typically cover more than a letter but less than a word. The size of the token dictionary could be, e.g., $D$ = 60,000.
An input token is then just mapped (i.e. embedded) into a vector $\in R^D$ that serves as the input to a stack of transformer layers.
Thus, to use pre-trained LLMs as autoregressive models for anything else but text, 
including for numerical time series data, a suitable mapping (and the reverse mapping) 
must be devised. 
To accomplish this, we convert the sequence of forecasting information sets, $\mathcal{I}_t = \{Y_{t-C:t}, \mathbf{X}_{t-C:t}^{(t)}, {X}^{(s)}\}$, 
into a sequence of representations suitable for the LLM. 
We map some (or all) of the $\mathcal{I}_t$ into the token embedding space through some parameterized continuous function, e.g., a linear or MLP-like function. 
Here, there will be no explicit tokens, but the dimension of the embedding space remains the same. 

We coarsely partition the Seq2seq models considered in this paper into three segments: an encoder block; a decoder block; and an output block. 
See Figure~\ref{fig:arch_tsfpt} for an illustration. 
The encoder block embeds the static and historic time series features; the decoder block decodes the hidden outputs of the encoder block; and the output block maps the decoder output to predictions in the shape of the target (optionally including time series future information). 

We consider several variations in each of these three categories.

%\vspace{-2mm}
%\subsubsection{Encoder Blocks}
\paragraph{Encoder Blocks.}
For encoder blocks, we consider the following.

\begin{itemize}[noitemsep,nolistsep,leftmargin=*]
\item
\textbf{\textit{CNN + MLP (MQCNN).}} 
MQCNN \citep{wen2018multihorizon} uses a WaveNet \citep{oord2016wavenet} CNN architecture to encode historic time series features into time-specific embeddings, and it uses a simple MLP to encode static time series features into time-agnostic embeddings. 
The encoder block output is a concatenation of these two embeddings and serves as a representation of the time series for the decoder.
\item \textbf{\textit{Multivariate Patching + Linear/MLP.}} We also use the Multivariate Patching strategy (described in Sec.~\ref{ssec:multipatch}) to aggregate information across historic time series and static features prior to a Linear or MLP embedding layer.
\end{itemize}

%\vspace{-2mm}
%\subsubsection{Decoder Blocks}
\paragraph{Decoder Blocks.}
For decoder blocks, we consider the following.

\begin{itemize}[noitemsep,nolistsep,leftmargin=*]
\item 
\textbf{\textit{MLP (MQCNN).}} 
When predicting 
for multiple horizons simultaneously, 
%the horizon mesh, 
the MQCNN decoder accounts for both ``local'' and ``global'' contexts with a series of horizon specific and horizon agnostic MLP layers \citep{wen2018multihorizon}.
\item
\textbf{\textit{Pre-trained LLM.}} 
We use three pre-trained LLMs as decoders to the embedded static and time series features: GPT-2 \citep{radford2019language}; Flan-T5 \cite{chung2022scaling}; and MPT-7B \cite{MosaicML2023Introducing}. 
While GPT-2 and MPT-7B are decoder-only models, Flan-T5 has both encoder and decoder transformer blocks. 
We experiment with using both the full Flan-T5 model as well as only the decoder of Flan-T5 as the pre-trained LLM. 
Each of these transformer blocks consists of multi-head attention, an MLP, and layer norms. 
As in TSFPT \citep{zhou2023fits}, we experiment with freezing the LLM and fine-tuning the layer norms, as well as freezing both LLM and layer~norms.
\item
\textbf{\textit{No Decoder.}} 
Since there are substantial differences between the structure of the MQCNN-based and LLM-based decoder blocks, we also look at a ``null'' decoder as a baseline, so that differences in the P50 and P90 quantile loss can be attributed to the addition of the MQCNN or LLM blocks.
\end{itemize}

%\vspace{-2mm}
%\subsubsection{Output Blocks}
\paragraph{Output Blocks.}
%\phantom{a}
For output blocks, we consider the following.

\begin{itemize}[noitemsep,nolistsep,leftmargin=*]
\item
The output of the pre-trained LLMs will correspond to output tokens to be converted to text, but we are rather interested in a multi-horizon time series forecast output. To that end, we use two output blocks to reshape the decoder block output to an appropriate size for our target: a simple linear layer; and a 2-layer~MLP. 
\end{itemize}

%\vspace{-1mm}
\subsection{Multivariate Patching for Time Series}
\label{ssec:multipatch}
%\vspace{-1mm}

``Patching'' \citep{nie2023time} has gained recent popularity in time-series forecasting as a pre-processing step for self-attention on univariate time series. 
Patching attempts to contextualize individual points in a time series by strided flattening of the time axis in windows of size $w$ and stride length $s$, so that each flattened window behaves as $w$ parallel features. 
In the case of forecasting, the initial values of the time series are repeated $s$ times to avoid temporal leakage prior to this process.

To understand its connection with self-attention, we can compare the process of self-attention on time series versus natural language. 
Generally, self-attention on raw time series data performs relatively poorly \citep{zeng2023transformers}; attending to individual points is often too granular to learn relevant temporal correlations across the series. 
Intuitively, each time point is similar to a letter in natural language data. 
Hence, patching provides a contextualized window of time points---similar to a word in natural language---providing a time-series analogue to semantic structure.

When originally formulated, patching was posed as an operation on a single time series \citep{nie2023time}. 
More recent implementations in the context of time series forecasting have remained operationally univariate \citep{zhou2023fits}. 
We propose a \emph{multivariate patching} procedure to pre-processes static and historical time series inputs for the LLM to ingest.  
Specifically, we define the multivariate patching block as follows:

{$\textbf{MultivariatePatching}(\mathbf{X}_{t-C:t}^{(t)}, X^{(s)}):$\\
%\vspace{-2mm}
\small
\begin{equation}
    \begin{aligned}
        \mathbf{X}_{t-C:t}^{(t,s)} &= \text{Concat}(\mathbf{X}_{t-C:t}^{(t)}, X^{(s)})\\
        \mathbf{X}_{t-C-s:t}^{(t,s)} &= \text{ZeroPad}(\mathbf{X}_{t-C:t}^{(t,s)})\\
        \mathbf{X}^{(p)} &= \text{Patch1d}(\mathbf{X}_{t-C-s:t}^{(t,s)})\\
        \widetilde{\mathbf{X}}^{(p)} &= \text{Flatten}(\mathbf{X}^{(p)})\\
        \mathbf{H}^{(p)} &= \text{Embed}(\widetilde{\mathbf{X}}^{(p)}).
    \end{aligned}
\end{equation}
}

\noindent
In this block, for product batch $B$, we first concatenate the static $B\times m$ feature matrix $X^{(s)}$ to the $B\times C \times d$ time series feature tensor $\mathbf{X}_{t-C:t}^{(t)}$ at each time point, resulting in an $B\times C \times (d + m)$ augmented time series tensor $\mathbf{X}_{t-C:t}^{(t,s)}$. 
We then pad the resulting tensor with zeros to avoid temporal leakage and reshape $\mathbf{X}_{t-C:t}^{(t,s)}$ into a $B\times p \times w(d+m)$ tensor $\mathbf{X}^{(p)}$, where $p$ is the number of patches, and $w$ is the window size of each patch. 
The tensor $\mathbf{X}^{(p)}$ is then flattened and passed into either a linear layer or two layer MLP as an embedding layer, outputting an $n\times p \times d_{\text{llm}}$ hidden representation tensor $\mathbf{H}^{(p)}$, where $d_{\text{llm}}$ is the hidden dimension of the LLM considered.

%\vspace{-1mm}
\subsection{Weight Analysis}
%\vspace{-1mm}

Finding the best architecture to use to embed time series covariates into a text embedding space is an ambiguous problem, and there is a need for NN model diagnostics (analogous to traditional regression diagnostics in traditional time series forecasting).
To assist in this architecture search, we use layer-level Empirical Spectral Densities (ESD) of weight matrices, using ideas from HTSR Theory~\citep{martin2019traditional,martin2020heavy,martin2021implicit}. 
Prior work has shown that aggregated shape metrics from layer ESDs can be used to predict the trends in the quality of SOTA CV and NLP models, even without access to training/testing data~\citep{martin2021predicting,yang2022evaluating}.
Our use here is analogous to the use of diagnostics for linear models or generalized models; and the methods that we use for time series forecasting that are based on HTSR Theory have been described recently~\citep{ww_timeseries_TR}.
%%in  two recent AMLC papers~\citep{wolfe-mahoney-amlc22,wolfe-mahoney-amlc23}.

In this paper, following recent work~\citep{ww_timeseries_TR}, we qualitatively and quantitatively assess the HT shape of the ESD, demonstrating that layer-level characteristics of the ESDs can be used to diagnose and predict model quality. % in our application.
In particular, we show that when layer-level ESDs do not exhibit a clear (T)PL shape, then the architecture may be sub-optimal for our forecasting task, and one should seek alternative architectures that yield better layer-level ESDs. 
Additionally, we show that when layer-level ESDs are approximately (T)PL, then existing HTSR metrics are strongly predictive of forecasting accuracy at both inter- and intra- model level, i.e., across different architectures and within the same architecture across different epochs.
\section{Main results}
\label{sxn:main_results}
%\vspace{-3mm}

%\michael{Idea1: take LLM and put MLP on each side and get almost SOTA.}

In this section, we describe our empirical setup and our main empirical results.

%\vspace{-1mm}
\subsection{Data}
\label{subsection:data}
%\vspace{-1mm}

The data we use for model training and evaluation consists of demand data from products sold nationwide by a large internet retailer.
The data contain a large number of historic time series, static, and future time series features. 
%Historic features include demand and indicators of customer interest (such as product views), in-stock rates, among other features. 
%third-party demand; Detailed Page Hits (DPH): total, in-stock and buybox; birth date, first received date, publication date, street date, product site launch date; 
%Future features include the distance to the nearest holiday for multiple holidays, binary features that show whether a holiday occurs in the upcoming week, as well as other ``causals,'' e.g., the number of days until the next known promotion date for a number of different type of promotions. 
%Static features include a number of catalog classifications (such as brand or category) as well as search keywords, list price, and additional static characteristics.
%\michael{Did we define GL anywhere.}
%
% NOTE: Changing this for punblic consumption
We use a small sample set of products that have relatively predicable demand, using a three year period as a training set and the following 52 week period as a test set.
%We use weekly demand series of 500,000 consistently high demand products from 2015 to 2019, where the three year period leading up to 2018 is used as training data and the 52 week period in 2018-2019 is used as testing data.
% We use weekly demand series of 500,000 consistently high demand products from 2015 to 2019, where the three year period leading up to 2018 is used as training data and the 52 week period in 2018-2019 is used as testing data.
%
% We use this relatively stable period to avoid strong COVID-19 related effects in the training or testing set, which may bias the results of the experiment.
%\michael{@malcolm: I assume that some of that par needs to be modified, would you do that?}\malcolm{@michael: I've modified this data section to mask a lot of information, but we'll see what the lawyers say.}

% \textbf{Weekly Retail Demand (2016-2020)} We use weekly demand series of around 50,000 sampled high-velocity products from different categories within the US marketplace from 3/27/2016 to 6/19/2019, where the three year period leading up to 2019 is used as training data and the 12 week period from 3/27/2019-6/19/2019 is used as testing data.

%\vspace{-1mm}
\subsection{Experiments}
%\vspace{-1mm}

We now describe the models we use in our continuous embedding experiments. 
Each of these models optimizes aggregate P50 and P90 quantile loss over the target periods, described in Sec.~\ref{subsection:data}.

\paragraph{Full Model Specifications.}

\begin{table}[H] %[h]
    \centering
    \scalebox{0.9}{
    \begin{tabular}{lrrr}
        \textbf{Model} & \textbf{Hidden Dimension} & \textbf{Total Params.} & \textbf{Trainable Params.} \\
        \midrule
        GPT-2 Small & 768 & 124MM & 37.2MM \\
        Flan-T5-small & 512 & 101MM & 24.7MM \\
        \phantom{Flan} \textit{Encoder Only} & 512 & 43.5MM & 24.7MM \\
        \phantom{Flan} \textit{Decoder Only} & 512 & 49.8MM & 24.7MM \\
        MPT-7B & 4096 & 6.8B & 196MM \\
        \midrule
        Linear Only & 768 & 37.2MM & 37.2MM \\
        MLP Only & 768 & 37.8MM & 37.8MM \\
        \bottomrule
    \end{tabular}}
    \vspace*{2mm}
    \caption{Parameter counts of fine-tuned LLMs and their effective ``trainable parameters'' in our model (e.g., layer norm weights and input/output blocks), as well as the ``Linear Only'' and ``MLP Only'' input/output block baselines.     %\malcolm{The ``encoder'' FFN on the flattened patched time series (Xs and Xt) makes the number of parameters very large.}
    }
    \label{tab:parameter_counts}
\end{table}

\begin{itemize}[noitemsep,nolistsep,leftmargin=*]
\item
\textbf{\textit{MQCNN (Baseline).}} 
As the SOTA model we use MQCNN~\cite{wen2018multihorizon} to represent a well-performing task-specific architecture for our forecasting problem.
In this model, historical time series inputs are embedded with a series of increasingly dilated causal convolutions, and a linear embedding is applied to static inputs. 
The decoder architecture contains MLPs of historical time series embeddings and static embeddings, in addition to future information. 
We evaluate the MQCNN model with and without the use of future information and 16-bit quantization. 
%\malcolm{We may need to remove the specificity in the architecture and just refer the readers to the public paper, which is less specific.}

\item
%\paragraph{\textit{Linear Only (Baseline).}} 
\textbf{\textit{Linear Only (Baseline).}} 
We benchmark the performance of a simple linear encoder and output layer. 
Specifically, after patching historical and static time series features, we use a linear map on each feature and time-point within the patch window as an encoder. The patched series is then expanded to the length of the original series, concatenated with available future information and decoded to the target sequence using a linear map.
We evaluate this model with and without the use of future information and 16-bit~quantization.
We use a hidden dimension of 768 for equivalence with the GPT-2 Small hidden dimension.

\item
\textbf{\textit{MLP Only (Baseline)}} 
We also benchmark the performance of simple two layer MLPs for the encoder and output layer. 
We again patch the historical and static time series features, and we use a MLP to embed the these features across each patch into an embedding dimension equal to the LLM. 
We use a second MLP to decode the embedded features into a multi-horizon prediction. 
We evaluate the model with and without the use of future features and with 16-bit quantization.
We again use a hidden dimension of 768 for equivalence with the GPT-2 Small hidden dimension.

\item
\textbf{Targeted Fine-Tuning of LLMs.} 
In our fine-tuning experiments, we use \textit{multivariate patching}, and pass each patch through a linear/MLP encoder layer to pre-trained LLMs, and pass the LLM output to a linear/MLP layer.
Following~\citep{zhou2023fits}, layers associated with the LLM (except layer norms) are frozen during training. 
We evaluate this model with and without future information, and with and without additionally frozen layer norms (e.g., a ``fully frozen'' LLM). 
We evaluate three LLM backbone models: GPT-2 Small, Flan-T5 Small, and MPT-7B.
We'll omit the ``Small'' suffix in what follows.
While GPT-2 uses a single Transformer stack, Flan-T5 uses a separate Transformer stack for the encoder and decoder; the encoder stack uses self-attention, while the decoder stack uses both self-attention and cross-attention. 
For Flan-T5, we evaluate performance when passing the linear embedding through the full model, only the encoder stack, and only the decoder stack. 
Finally, for a more recent LLM benchmark, we fine-tune the 7 billion parameter model MPT-7B--- an open source model developed by MosaicML in 2023 which shows competitive performance with Llama-7B on a range of benchmarks~\cite{MosaicML2023Introducing}.
\end{itemize}

\noindent
Table \ref{tab:parameter_counts} shows the LLMs used in this work and their parameter counts, as well as the parameter counts of the baselines. 
While these models have a large number of parameters, the number of their layer norm parameters are comparatively low.

%\vspace{-1mm}
\subsection{Summary of Results}
%\vspace{-1mm}
{\small
\begin{table}[h]
    \centering
    \scalebox{.8}{
    \begin{tabular}{lccccc}
         & \multicolumn{2}{c}{\textbf{Training Permutations}} &  & \multicolumn{2}{c}{\textbf{Quantile Weighted Error}} \\ 
         \cmidrule(lr){2-3} \cmidrule(lr){5-6}
        \textbf{Architecture} & \textit{16-bit} & \textit{+Future Info.} & Epochs & P50 & P90 \\
        \midrule
        {\large \textit{\textbf{Baselines}}} &  & &  &  &  \\
        \midrule
         MQCNN & & \checkmark & 100 & \textbf{0.996} & 1.007 \\
         & \checkmark & & 100 & 1.016 & 1.051 \\
         & \checkmark & \checkmark & 100 & 1.000 & \textbf{1.000} \\
        \cmidrule(lr){2-6}
        Linear Only & \checkmark & & 100 & 1.183 & 1.372 \\
        & \checkmark & \checkmark & 100 & 1.181 & 1.367 \\
        \cmidrule(lr){2-6}
        MLP Only & \checkmark & \checkmark & 100 & 1.116 & 1.248 \\
        \toprule
        {\large \textit{\textbf{Linear Adapter}}} &  & &  &  &  \\
        \toprule
        GPT-2 Linear & \checkmark & & 100 & 1.157 & 1.386 \\
        & \checkmark & \checkmark & 100 & 1.136 & 1.362 \\
        \phantom{GPT-2} \textit{Fully Frozen} & \checkmark & \checkmark & 100 & 1.164 & 1.403 \\
        \cmidrule(lr){2-6}
        Flan-T5 Linear & \checkmark & & 100 & 1.116 & 1.282 \\
         & \checkmark & \checkmark & 100 & 1.051 & 1.166 \\
         \phantom{Flan-T5} \textit{Fully Frozen} & \checkmark & \checkmark & 100 & 1.114 & 1.280 \\
         \phantom{Flan-T5} \textit{{Encoder Only}} & \checkmark & \checkmark & 100 & 1.068 & 1.198 \\
        \phantom{Flan-T5} \textit{{Decoder Only}} & \checkmark & \checkmark & 100 & 1.049 & 1.156 \\
        \cmidrule(lr){2-6}
        MPT-7B Linear & \checkmark & \checkmark & 10 & \textcolor{violet}{\textbf{1.005}} & \textcolor{violet}{\textbf{1.029}} \\
        \toprule
        {\large \textit{\textbf{MLP Adapter}}} &  & &  &  &  \\
        \toprule
        GPT-2 MLP & \checkmark & & 100 & 1.033 & 1.130 \\
        & \checkmark & \checkmark & 100 & 1.000 & 1.032 \\
        \cmidrule(lr){2-6}
        Flan-T5 MLP & \checkmark & \checkmark & 100 & 1.004 & 1.039 \\
         \phantom{Flan-T5} \textit{{Decoder Only}} & \checkmark & \checkmark & 100 & 0.996 & 1.028 \\
        \cmidrule(lr){2-6}        
        MPT-7B MLP & \checkmark & \checkmark & 10 & \textcolor{red}{\textbf{0.994}} & \textcolor{red}{\textbf{1.005}} \\
        \bottomrule
    \end{tabular}}
    \vspace*{2mm}
    \caption{P50 and P90 quantile weighted errors for MQCNN and the fine-tuned LLMs on the 52-week test period.}
    \label{tab:500k}
\end{table}
}

Empirical evaluation of our results the product set is summarized in Table~\ref{tab:500k}. 
For each experiment, we use a patch window size of 12 and stride of 6. 
We display P50 and P90 quantile weighted errors after 100 epochs.%
\footnote{MPT-7B is only trained for 10 epochs due to computational constraints.}
Several summary conclusions can be drawn.
\begin{itemize}
\item
First, we find clear evidence that LLMs pre-trained \textit{only on language tasks} contain relevant information for multivariate time series forecasting. 
Namely, across all specifications, all FPT (frozen pre-trained) variations outperform the ``Linear Only'' baseline.
This is true even when FPT is \textit{fully frozen}.
Hence, this effect is not straightforwardly caused by a larger number of learned parameters (i.e., layer norms). 
\item 
Second, we find that \textit{not all LLMs are created equal}.
In particular, a pre-trained Flan-T5 is evidently more suited for time series forecasting than is a pre-trained GPT-2, and MPT-7B even moreso. 
%\michael{Text here also needs to be modified as we ass new models.}
\item 
Finally, \textit{pre-trained LLMs are close to SOTA}, and MPT-7B outperforms even MQCNN on P50 quantile loss.
%\michael{Can we be a bit more precise / detailed / strong than ``reasonably close''?  Even ``comparable''?}
The best LLM linear embedding model \textit{improves} quantile weighted error by 3\% on P50 and degrades by 3\% on P90. 
When an MLP embedding is used on MPT-7B, it slightly improves over MQCNN on both P50 and P90 quantile loss.
\end{itemize}

\begin{figure}[h]
  \centering
  \begin{subfigure}{0.33\textwidth}
    \centering
    \includegraphics[width=\textwidth]{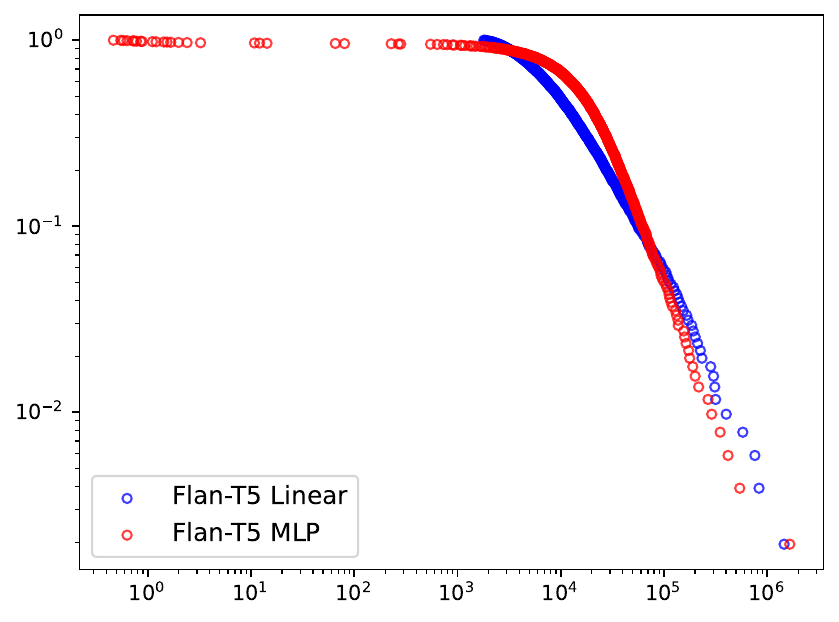}
  \end{subfigure}%
  \begin{subfigure}{0.33\textwidth}
    \centering
    \includegraphics[width=\textwidth]{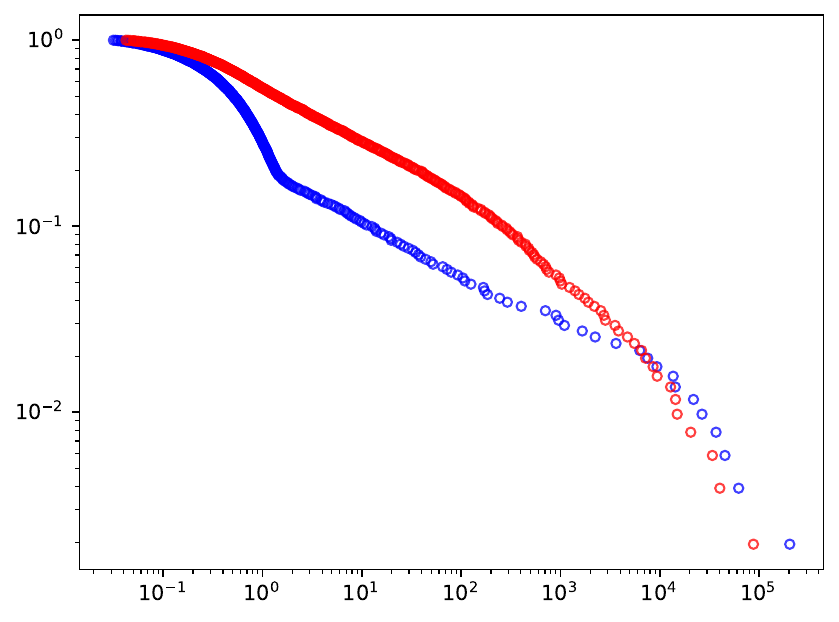}
  \end{subfigure}%
  \begin{subfigure}{0.33\textwidth}
    \centering
    \includegraphics[width=\textwidth]{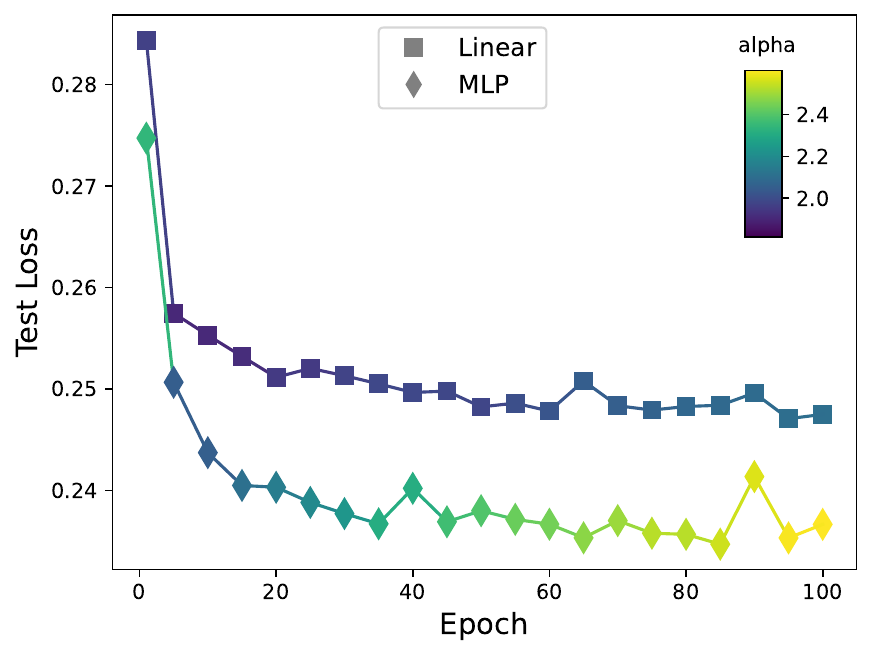}
  \end{subfigure}
  \caption{{\bf Layer-level weight analysis identifies sub-optimal model architecture and predicts forecasting accuracy.} Complementary cumulative distribution function (CCDF) of weight matrix spectrum for the first embedding layer (left) and the output layer (middle). %when the embedding/decoding layer is linear and MLP, respectively. %Good models should have ESDs that demonstrate power-law (PL) tails, e.g., when the trail CCDF of ESD is approximately linear on log-log scale. 
  %With linear decoding layer, the CCDF for the output layer has a strange convex kink and does not result in PL tail. MLP decoding layer fixes this as the corresponding CCDF tends to closely follow a (T)PL shape. For the embedding layer, both CCDFs demonstrate PL tails, but {\em Flan-T5 MLP} has a steeper slope (i.e. better $\alpha$ metric) than {\em Flan-T5 Linear}. 
  The rightmost plot shows evolution of P50 quantile test loss across epochs for both architectures, where markers and lines are colored according to the fitted $\alpha$.
  }
  \label{fig:weight_analysis}
\end{figure}

To complement the results summarized in Table~\ref{tab:500k}, we use HTSR Theory to perform model quality diagnostics by examining layer-level ESDs. 
Specifically, HTSR theory suggests that the degree to which the eigenvalues of the layer weight Gram matrices represent a Power Law (PL) distribution is correlated with model quality and generalization \cite[e.g.,][]{martin2019traditional, martin2020heavy, martin2021implicit, martin2021predicting}.
An example result in shown in Figure~\ref{fig:weight_analysis}.
The two left plots show the empirical complementary cumulative distribution function (CCDF), with \textit{x}-axis the magnitude of the Gram matrix eigenvalue and \textit{y}-axis equal to $1-\widehat{F}(x)$, the empirical CCDF value.
The right plot shows a summary statistic of a fitted PL distribution to the Gram matrix eigenvalues relative to test loss across training epochs.
Additional results are shown in Appendix~\ref{appendix:htsr}
\michael{I made some changes below, reordering and rewording, e.g. ``{\em Flan-T5 Linear}'' there.  Would you go through all the bullets and put names corresponding to what we use the tables to help the reader?  Would you also check that I did it correctly, there and elsewhere in the bullets?  We should probably make sure we use the names very pedantically and consistently throughout.}
\begin{itemize}
\item 
The left plot in Figure~\ref{fig:weight_analysis} shows that the ESD of {\em Flan-T5 MLP} has a steeper CCDF---and hence a lower/better value of the $\alpha$ metric---than that of {\em Flan-T5 Linear}.
This corresponds to the higher quality of {\em Flan-T5 MLP}.
\item 
The middle plot in Figure~\ref{fig:weight_analysis} shows that {\em Flan-T5 Linear}, i.e., the LLM (Flan-T5) model with linear decoding, exhibits an exotic/unusual ESD in its output layer, having a convex kink that is clearly not PL and that is not typically seen in SOTA models~\citep{ww_timeseries_TR}.
On the other hand, {\em Flan-T5 MLP}, i.e., the LLM (Flan-T5) with MLP decoding, fixes the problem and results in an ESD that much more closely follows (T)PL. 
\item
In the rightmost plot of Figure~\ref{fig:weight_analysis}, we color the test loss curve according to the fitted $\alpha$ metric. 
Consistent with prior work~\citep{ww_timeseries_TR}, the $\alpha$ metric is strongly predictive of model quality, with better values being smaller and closer to $2$, at both intra- and inter-model level. 
We provide additional details and diagnostics in the appendix, including a comparison between different LLMs (see Sec.~\ref{sec:esd_gpt2_vs_flant5}) and a comparison with linear baselines (see Sec.~\ref{sec:esd_fpt_vs_non_fpt}).
\end{itemize}

\section{Discussion and Future Work}
\label{sxn:discussion}
%\vspace{-2mm}

We consider whether an LLM pretrained for language prediction is able to transfer its ``knowledge'' to time-series forecasting, a seemingly very different prediction domain. 
To do so, we look at a SOTA 
%%Amazon ASIN 
demand forecasting problems. 
Our empirical results suggest that the answer is a (preliminary, but definitive) ``yes.'' 
While work has shown that pre-trained LLMs may not outperform their attention-based counterparts trained from scratch \cite{tan2024languagemodelsactuallyuseful}, we find that relative to our ``Linear Only'' and ``MLP Only'' baselines with an equal or greater number of trainable parameters (see Table \ref{tab:parameter_counts}), pre-trained LLM inclusion does improve performance for our data.
Importantly, beyond our initial HTSR-based diagnostic evaluation, we have not ventured into trying to identify what exactly it is that the LLM is able to transfer to time-series prediction. 
We refer the reader to~\citep{zhou2023fits} for initial speculations. 
We recognize a number of additional limitations of this work.
First, training and test accuracy are computed on limited dataset of products.
Because of this, higher parameter models may be able to ``overfit'' to the population.
That being said, the results indicate that the models can at least generalize to the unseen 52-week period following training.
Moreover, the data are made up of the fastest products, which may be significantly easier to predict than time series in general.
Second, due compute limitations, we are only able to fine-tune for 100 epochs for most LLMs, and 10 epochs for MPT-7B; the latter could observe even better performance over a 100 epoch training period, or show evidence of overfitting.
Finally, we are limited to the open-source models listed in Table \ref{tab:500k}, which are significantly behind many proprietary SOTA LLMs.
%%% I THINK STALE FROM AMLCW % Of the two approaches for mapping the time series data to the embedding space of an LLM, the continuous mapping (as opposed to the discrete mapping, which involved quantization and tokenization) was clearly better. 
%%% I THINK STALE FROM AMLCW % We were able to reach SOTA forecasting accuracies with this approach. 
%%
Apart from bridging these gaps, our initial study leaves additional room for future work. 
%%% ALSO STALE
%For example, our quantization approach now discards all but the main time-series. 
%We intend to experiment with Vector Quantization to allow multivariate input.  
One motivation for using a ``foundation model'' approach is to be able to adapt to a new task (such as forecasting for a new product, or starting to forecast for a new marketplace) with very little or no data from the new task. 
We intend to run a comprehensive set of experiments to characterize this ability.

% Uncomment to see all outstanding todos.
% Leave commented since it introduces some latex errors
% \listoftodos

%\newpage

%\bibliographystyle{apalike}
%\bibliography{amlc23-bib}
%
\bibliographystyle{unsrt}
\bibliography{main.bib}

\begin{thebibliography}{10}

\bibitem{harvey90}
Andrew~C. Harvey.
\newblock {\em The Econometric Analysis of Time Series}.
\newblock MIT Press, 1990.

\bibitem{wen2018multihorizon}
Ruofeng Wen, Kari Torkkola, Balakrishnan Narayanaswamy, and Dhruv Madeka.
\newblock A multi-horizon quantile recurrent forecaster, 2018.

\bibitem{eisenach2020mqtransformer}
Carson Eisenach, Yagna Patel, and Dhruv Madeka.
\newblock Mqtransformer: Multi-horizon forecasts with context dependent and
  feedback-aware attention.
\newblock {\em arXiv preprint arXiv:2009.14799}, 2020.

\bibitem{gong2023multimodal}
Tao Gong, Chengqi Lyu, Shilong Zhang, Yudong Wang, Miao Zheng, Qian Zhao,
  Kuikun Liu, Wenwei Zhang, Ping Luo, and Kai Chen.
\newblock Multimodal-gpt: A vision and language model for dialogue with humans.
\newblock {\em arXiv preprint arXiv:2305.04790}, 2023.

\bibitem{gao2023llama}
Peng Gao, Jiaming Han, Renrui Zhang, Ziyi Lin, Shijie Geng, Aojun Zhou, Wei
  Zhang, Pan Lu, Conghui He, Xiangyu Yue, et~al.
\newblock Llama-adapter v2: Parameter-efficient visual instruction model.
\newblock {\em arXiv preprint arXiv:2304.15010}, 2023.

\bibitem{hahn2023theory}
Michael Hahn and Navin Goyal.
\newblock A theory of emergent in-context learning as implicit structure
  induction.
\newblock {\em arXiv preprint arXiv:2303.07971}, 2023.

\bibitem{hagendorff2023machine}
Thilo Hagendorff.
\newblock Machine psychology: Investigating emergent capabilities and behavior
  in large language models using psychological methods.
\newblock {\em arXiv preprint arXiv:2303.13988}, 2023.

\bibitem{foundationStanford_TR}
Rishi Bommasani et~al.
\newblock On the opportunities and risks of foundation models.
\newblock {\em arXiv preprint arXiv:2108.07258}, 2021.

\bibitem{sung2022vl}
Yi-Lin Sung, Jaemin Cho, and Mohit Bansal.
\newblock Vl-adapter: Parameter-efficient transfer learning for
  vision-and-language tasks.
\newblock In {\em Proceedings of the IEEE/CVF Conference on Computer Vision and
  Pattern Recognition}, pages 5227--5237, 2022.

\bibitem{chen2022simple}
Yutong Chen, Fangyun Wei, Xiao Sun, Zhirong Wu, and Stephen Lin.
\newblock A simple multi-modality transfer learning baseline for sign language
  translation.
\newblock In {\em Proceedings of the IEEE/CVF Conference on Computer Vision and
  Pattern Recognition}, pages 5120--5130, 2022.

\bibitem{zhu2023minigpt}
Deyao Zhu, Jun Chen, Xiaoqian Shen, Xiang Li, and Mohamed Elhoseiny.
\newblock Minigpt-4: Enhancing vision-language understanding with advanced
  large language models.
\newblock {\em arXiv preprint arXiv:2304.10592}, 2023.

\bibitem{zeng2022socratic}
Andy Zeng, Maria Attarian, Brian Ichter, Krzysztof Choromanski, Adrian Wong,
  Stefan Welker, Federico Tombari, Aveek Purohit, Michael Ryoo, Vikas
  Sindhwani, et~al.
\newblock Socratic models: Composing zero-shot multimodal reasoning with
  language.
\newblock {\em arXiv preprint arXiv:2204.00598}, 2022.

\bibitem{zhang2023multimodal}
Zhuosheng Zhang, Aston Zhang, Mu~Li, Hai Zhao, George Karypis, and Alex Smola.
\newblock Multimodal chain-of-thought reasoning in language models.
\newblock {\em arXiv preprint arXiv:2302.00923}, 2023.

\bibitem{liu2023visual}
Haotian Liu, Chunyuan Li, Qingyang Wu, and Yong~Jae Lee.
\newblock Visual instruction tuning.
\newblock {\em arXiv preprint arXiv:2304.08485}, 2023.

\bibitem{zeng2023transformers}
Ailing Zeng, Muxi Chen, Lei Zhang, and Qiang Xu.
\newblock Are transformers effective for time series forecasting?
\newblock In {\em Proceedings of the AAAI conference on artificial
  intelligence}, volume~37, pages 11121--11128, 2023.

\bibitem{kitaev2020reformer}
Nikita Kitaev, Łukasz Kaiser, and Anselm Levskaya.
\newblock Reformer: The efficient transformer, 2020.

\bibitem{zhou2021informer}
Haoyi Zhou, Shanghang Zhang, Jieqi Peng, Shuai Zhang, Jianxin Li, Hui Xiong,
  and Wancai Zhang.
\newblock Informer: Beyond efficient transformer for long sequence time-series
  forecasting, 2021.

\bibitem{wu2022autoformer}
Haixu Wu, Jiehui Xu, Jianmin Wang, and Mingsheng Long.
\newblock Autoformer: Decomposition transformers with auto-correlation for
  long-term series forecasting, 2022.

\bibitem{zhou2022fedformer}
Tian Zhou, Ziqing Ma, Qingsong Wen, Xue Wang, Liang Sun, and Rong Jin.
\newblock {FEDformer}: Frequency enhanced decomposed transformer for long-term
  series forecasting, 2022.

\bibitem{zhou2023fits}
Tian Zhou, PeiSong Niu, Xue Wang, Liang Sun, and Rong Jin.
\newblock One fits all: Power general time series analysis by pretrained {LM},
  2023.

\bibitem{chronos_TR}
Abdul~Fatir Ansari, Lorenzo Stella, Caner Turkmen, Xiyuan Zhang, Pedro Mercado,
  Huibin Shen, Oleksandr Shchur, Syama~Sundar Rangapuram, Sebastian~Pineda
  Arango, Shubham Kapoor, Jasper Zschiegner, Danielle~C. Maddix, Hao Wang,
  Michael~W. Mahoney, Kari Torkkola, Andrew~Gordon Wilson, Michael
  Bohlke-Schneider, and Yuyang Wang.
\newblock Chronos: Learning the language of time series.
\newblock {\em arXiv preprint arXiv:2403.07815}, 2024.

\bibitem{martin2019traditional}
Charles~H Martin and Michael~W Mahoney.
\newblock Traditional and heavy tailed self regularization in neural network
  models.
\newblock In {\em International Conference on Machine Learning}, pages
  4284--4293. PMLR, 2019.

\bibitem{martin2020heavy}
Charles~H Martin and Michael~W Mahoney.
\newblock Heavy-tailed universality predicts trends in test accuracies for very
  large pre-trained deep neural networks.
\newblock In {\em Proceedings of the 2020 SIAM International Conference on Data
  Mining}, pages 505--513. SIAM, 2020.

\bibitem{martin2021implicit}
Charles~H Martin and Michael~W Mahoney.
\newblock Implicit self-regularization in deep neural networks: Evidence from
  random matrix theory and implications for learning.
\newblock {\em Journal of Machine Learning Research}, 22(165):1--73, 2021.

\bibitem{martin2021predicting}
Charles~H Martin, Tongsu~Serena Peng, and Michael~W Mahoney.
\newblock Predicting trends in the quality of state-of-the-art neural networks
  without access to training or testing data.
\newblock {\em Nature Communications}, 12(1):1--13, 2021.

\bibitem{yang2022evaluating}
Yaoqing Yang, Ryan Theisen, Liam Hodgkinson, Joseph~E Gonzalez, Kannan
  Ramchandran, Charles~H Martin, and Michael~W Mahoney.
\newblock Evaluating natural language processing models with generalization
  metrics that do not need access to any training or testing data.
\newblock {\em arXiv preprint arXiv:2202.02842}, 2022.

\bibitem{vaswani2017attention}
Ashish Vaswani, Noam Shazeer, Niki Parmar, Jakob Uszkoreit, Llion Jones,
  Aidan~N Gomez, {\L}ukasz Kaiser, and Illia Polosukhin.
\newblock Attention is all you need.
\newblock {\em Advances in neural information processing systems}, 30, 2017.

\bibitem{devlin2019bert}
Jacob Devlin, Ming-Wei Chang, Kenton Lee, and Kristina Toutanova.
\newblock {BERT}: Pre-training of deep bidirectional transformers for language
  understanding, 2019.

\bibitem{radford2019language}
Alec Radford, Jeffrey Wu, Rewon Child, David Luan, Dario Amodei, Ilya
  Sutskever, et~al.
\newblock Language models are unsupervised multitask learners.
\newblock {\em OpenAI blog}, 1(8):9, 2019.

\bibitem{brown2020language}
Tom Brown, Benjamin Mann, Nick Ryder, Melanie Subbiah, Jared~D Kaplan, Prafulla
  Dhariwal, Arvind Neelakantan, Pranav Shyam, Girish Sastry, Amanda Askell,
  et~al.
\newblock Language models are few-shot learners.
\newblock {\em Advances in neural information processing systems},
  33:1877--1901, 2020.

\bibitem{raffel2020exploring}
Colin Raffel, Noam Shazeer, Adam Roberts, Katherine Lee, Sharan Narang, Michael
  Matena, Yanqi Zhou, Wei Li, and Peter~J Liu.
\newblock Exploring the limits of transfer learning with a unified text-to-text
  transformer.
\newblock {\em The Journal of Machine Learning Research}, 21(1):5485--5551,
  2020.

\bibitem{chung2022scaling}
Hyung~Won Chung, Le~Hou, Shayne Longpre, Barret Zoph, Yi~Tay, William Fedus,
  Eric Li, Xuezhi Wang, Mostafa Dehghani, Siddhartha Brahma, et~al.
\newblock Scaling instruction-finetuned language models.
\newblock {\em arXiv preprint arXiv:2210.11416}, 2022.

\bibitem{MosaicML2023Introducing}
MosaicML~NLP Team.
\newblock Introducing {MPT-7B}: A new standard for open-source, commercially
  usable llms, 2023.
\newblock Accessed: 2023-05-05.

\bibitem{wen2022transformers}
Qingsong Wen, Tian Zhou, Chaoli Zhang, Weiqi Chen, Ziqing Ma, Junchi Yan, and
  Liang Sun.
\newblock Transformers in time series: A survey.
\newblock {\em arXiv preprint arXiv:2202.07125}, 2022.

\bibitem{wu2021autoformer}
Haixu Wu, Jiehui Xu, Jianmin Wang, and Mingsheng Long.
\newblock Autoformer: Decomposition transformers with auto-correlation for
  long-term series forecasting.
\newblock {\em Advances in Neural Information Processing Systems},
  34:22419--22430, 2021.

\bibitem{nie2023time}
Yuqi Nie, Nam~H. Nguyen, Phanwadee Sinthong, and Jayant Kalagnanam.
\newblock A time series is worth 64 words: Long-term forecasting with
  transformers, 2023.

\bibitem{yu2023temporal}
Xinli Yu, Zheng Chen, Yuan Ling, Shujing Dong, Zongyi Liu, and Yanbin Lu.
\newblock Temporal data meets {LLM} -- explainable financial time series
  forecasting, 2023.

\bibitem{fatir2024ChronosBolt}
Abdul Fatir~Ansari et~al.
\newblock {Chronos: Learning the Language of Time Series (Github)}.
\newblock
  \url{https://github.com/amazon-science/chronos-forecasting?tab=readme-ov-file#zero-shot-results},
  2024.

\bibitem{NIPS2017_a96b65a7}
Yuanzhi Li and Yang Yuan.
\newblock Convergence analysis of two-layer neural networks with relu
  activation.
\newblock In I.~Guyon, U.~Von Luxburg, S.~Bengio, H.~Wallach, R.~Fergus,
  S.~Vishwanathan, and R.~Garnett, editors, {\em Advances in Neural Information
  Processing Systems}, volume~30. Curran Associates, Inc., 2017.

\bibitem{frankle2020early}
Jonathan Frankle, David~J Schwab, and Ari~S Morcos.
\newblock The early phase of neural network training.
\newblock {\em arXiv preprint arXiv:2002.10365}, 2020.

\bibitem{damian2022neural}
Alexandru Damian, Jason Lee, and Mahdi Soltanolkotabi.
\newblock Neural networks can learn representations with gradient descent.
\newblock In {\em Conference on Learning Theory}, pages 5413--5452. PMLR, 2022.

\bibitem{ww_timeseries_TR}
Malcolm~L. Wolff and Michael~W. Mahoney.
\newblock Improved weight matrix diagnostics for time series forecasting
  models.
\newblock {\em arXiv preprint arXiv:2400.00000}, 2024.

\bibitem{oord2016wavenet}
Aaron van~den Oord, Sander Dieleman, Heiga Zen, Karen Simonyan, Oriol Vinyals,
  Alex Graves, Nal Kalchbrenner, Andrew Senior, and Koray Kavukcuoglu.
\newblock Wavenet: A generative model for raw audio.
\newblock {\em arXiv preprint arXiv:1609.03499}, 2016.

\bibitem{wei2022emergent}
Jason Wei, Yi~Tay, Rishi Bommasani, Colin Raffel, Barret Zoph, Sebastian
  Borgeaud, Dani Yogatama, Maarten Bosma, Denny Zhou, Donald Metzler, Ed~H.
  Chi, Tatsunori Hashimoto, Oriol Vinyals, Percy Liang, Jeff Dean, and William
  Fedus.
\newblock Emergent abilities of large language models, 2022.

\bibitem{chowdhery2022palm}
Aakanksha Chowdhery, Sharan Narang, Jacob Devlin, Maarten Bosma, Gaurav Mishra,
  Adam Roberts, Paul Barham, Hyung~Won Chung, Charles Sutton, Sebastian
  Gehrmann, et~al.
\newblock Palm: Scaling language modeling with pathways.
\newblock {\em arXiv preprint arXiv:2204.02311}, 2022.

\bibitem{openai2023gpt4}
OpenAI, :, Josh Achiam, Steven Adler, Sandhini Agarwal, Lama Ahmad, Ilge
  Akkaya, Florencia~Leoni Aleman, Diogo Almeida, Janko Altenschmidt, Sam
  Altman, Shyamal Anadkat, Red Avila, Igor Babuschkin, Suchir Balaji, Valerie
  Balcom, Paul Baltescu, Haiming Bao, Mo~Bavarian, Jeff Belgum, Irwan Bello,
  Jake Berdine, Gabriel Bernadett-Shapiro, Christopher Berner, Lenny Bogdonoff,
  Oleg Boiko, Madelaine Boyd, Anna-Luisa Brakman, Greg Brockman, Tim Brooks,
  Miles Brundage, Kevin Button, Trevor Cai, Rosie Campbell, Andrew Cann,
  Brittany Carey, Chelsea Carlson, Rory Carmichael, Brooke Chan, Che Chang,
  Fotis Chantzis, Derek Chen, Sully Chen, Ruby Chen, Jason Chen, Mark Chen, Ben
  Chess, Chester Cho, Casey Chu, Hyung~Won Chung, Dave Cummings, Jeremiah
  Currier, Yunxing Dai, Cory Decareaux, Thomas Degry, Noah Deutsch, Damien
  Deville, Arka Dhar, David Dohan, Steve Dowling, Sheila Dunning, Adrien
  Ecoffet, Atty Eleti, Tyna Eloundou, David Farhi, Liam Fedus, Niko Felix,
  Simón~Posada Fishman, Juston Forte, Isabella Fulford, Leo Gao, Elie Georges,
  Christian Gibson, Vik Goel, Tarun Gogineni, Gabriel Goh, Rapha Gontijo-Lopes,
  Jonathan Gordon, Morgan Grafstein, Scott Gray, Ryan Greene, Joshua Gross,
  Shixiang~Shane Gu, Yufei Guo, Chris Hallacy, Jesse Han, Jeff Harris, Yuchen
  He, Mike Heaton, Johannes Heidecke, Chris Hesse, Alan Hickey, Wade Hickey,
  Peter Hoeschele, Brandon Houghton, Kenny Hsu, Shengli Hu, Xin Hu, Joost
  Huizinga, Shantanu Jain, Shawn Jain, Joanne Jang, Angela Jiang, Roger Jiang,
  Haozhun Jin, Denny Jin, Shino Jomoto, Billie Jonn, Heewoo Jun, Tomer Kaftan,
  Łukasz Kaiser, Ali Kamali, Ingmar Kanitscheider, Nitish~Shirish Keskar,
  Tabarak Khan, Logan Kilpatrick, Jong~Wook Kim, Christina Kim, Yongjik Kim,
  Hendrik Kirchner, Jamie Kiros, Matt Knight, Daniel Kokotajlo, Łukasz
  Kondraciuk, Andrew Kondrich, Aris Konstantinidis, Kyle Kosic, Gretchen
  Krueger, Vishal Kuo, Michael Lampe, Ikai Lan, Teddy Lee, Jan Leike, Jade
  Leung, Daniel Levy, Chak~Ming Li, Rachel Lim, Molly Lin, Stephanie Lin,
  Mateusz Litwin, Theresa Lopez, Ryan Lowe, Patricia Lue, Anna Makanju, Kim
  Malfacini, Sam Manning, Todor Markov, Yaniv Markovski, Bianca Martin, Katie
  Mayer, Andrew Mayne, Bob McGrew, Scott~Mayer McKinney, Christine McLeavey,
  Paul McMillan, Jake McNeil, David Medina, Aalok Mehta, Jacob Menick, Luke
  Metz, Andrey Mishchenko, Pamela Mishkin, Vinnie Monaco, Evan Morikawa, Daniel
  Mossing, Tong Mu, Mira Murati, Oleg Murk, David Mély, Ashvin Nair, Reiichiro
  Nakano, Rajeev Nayak, Arvind Neelakantan, Richard Ngo, Hyeonwoo Noh, Long
  Ouyang, Cullen O'Keefe, Jakub Pachocki, Alex Paino, Joe Palermo, Ashley
  Pantuliano, Giambattista Parascandolo, Joel Parish, Emy Parparita, Alex
  Passos, Mikhail Pavlov, Andrew Peng, Adam Perelman, Filipe de~Avila
  Belbute~Peres, Michael Petrov, Henrique~Ponde de~Oliveira~Pinto, Michael,
  Pokorny, Michelle Pokrass, Vitchyr Pong, Tolly Powell, Alethea Power, Boris
  Power, Elizabeth Proehl, Raul Puri, Alec Radford, Jack Rae, Aditya Ramesh,
  Cameron Raymond, Francis Real, Kendra Rimbach, Carl Ross, Bob Rotsted, Henri
  Roussez, Nick Ryder, Mario Saltarelli, Ted Sanders, Shibani Santurkar, Girish
  Sastry, Heather Schmidt, David Schnurr, John Schulman, Daniel Selsam, Kyla
  Sheppard, Toki Sherbakov, Jessica Shieh, Sarah Shoker, Pranav Shyam, Szymon
  Sidor, Eric Sigler, Maddie Simens, Jordan Sitkin, Katarina Slama, Ian Sohl,
  Benjamin Sokolowsky, Yang Song, Natalie Staudacher, Felipe~Petroski Such,
  Natalie Summers, Ilya Sutskever, Jie Tang, Nikolas Tezak, Madeleine Thompson,
  Phil Tillet, Amin Tootoonchian, Elizabeth Tseng, Preston Tuggle, Nick Turley,
  Jerry Tworek, Juan Felipe~Cerón Uribe, Andrea Vallone, Arun Vijayvergiya,
  Chelsea Voss, Carroll Wainwright, Justin~Jay Wang, Alvin Wang, Ben Wang,
  Jonathan Ward, Jason Wei, CJ~Weinmann, Akila Welihinda, Peter Welinder, Jiayi
  Weng, Lilian Weng, Matt Wiethoff, Dave Willner, Clemens Winter, Samuel
  Wolrich, Hannah Wong, Lauren Workman, Sherwin Wu, Jeff Wu, Michael Wu, Kai
  Xiao, Tao Xu, Sarah Yoo, Kevin Yu, Qiming Yuan, Wojciech Zaremba, Rowan
  Zellers, Chong Zhang, Marvin Zhang, Shengjia Zhao, Tianhao Zheng, Juntang
  Zhuang, William Zhuk, and Barret Zoph.
\newblock {GPT-4 Technical Report}, 2023.

\bibitem{oquab2023dinov2}
Maxime Oquab, Timothée Darcet, Théo Moutakanni, Huy Vo, Marc Szafraniec,
  Vasil Khalidov, Pierre Fernandez, Daniel Haziza, Francisco Massa, Alaaeldin
  El-Nouby, Mahmoud Assran, Nicolas Ballas, Wojciech Galuba, Russell Howes,
  Po-Yao Huang, Shang-Wen Li, Ishan Misra, Michael Rabbat, Vasu Sharma, Gabriel
  Synnaeve, Hu~Xu, Hervé Jegou, Julien Mairal, Patrick Labatut, Armand Joulin,
  and Piotr Bojanowski.
\newblock {DINOv2}: Learning robust visual features without supervision, 2023.

\bibitem{chang2024llm4ts}
Ching Chang, Wei-Yao Wang, Wen-Chih Peng, and Tien-Fu Chen.
\newblock {LLM4TS}: Aligning pre-trained llms as data-efficient time-series
  forecasters, 2024.

\bibitem{jin2024timellm}
Ming Jin, Shiyu Wang, Lintao Ma, Zhixuan Chu, James~Y. Zhang, Xiaoming Shi,
  Pin-Yu Chen, Yuxuan Liang, Yuan-Fang Li, Shirui Pan, and Qingsong Wen.
\newblock {Time-LLM}: Time series forecasting by reprogramming large language
  models, 2024.

\bibitem{tan2024languagemodelsactuallyuseful}
Mingtian Tan, Mike~A. Merrill, Vinayak Gupta, Tim Althoff, and Thomas
  Hartvigsen.
\newblock Are language models actually useful for time series forecasting?,
  2024.

\end{thebibliography}

%%%%%%%%%%%%%%%%%%%%%%%%%%%%%%%%%%%%%%%%%%%%%%%%%%%%%%%%%%%%%%%%%%%%%%%%%%%%%%%
%%%%%%%%%%%%%%%%%%%%%%%%%%%%%%%%%%%%%%%%%%%%%%%%%%%%%%%%%%%%%%%%%%%%%%%%%%%%%%%
% APPENDIX
%%%%%%%%%%%%%%%%%%%%%%%%%%%%%%%%%%%%%%%%%%%%%%%%%%%%%%%%%%%%%%%%%%%%%%%%%%%%%%%
%%%%%%%%%%%%%%%%%%%%%%%%%%%%%%%%%%%%%%%%%%%%%%%%%%%%%%%%%%%%%%%%%%%%%%%%%%%%%%%

%\newpage

\appendix
%\onecolumn

%\newpage
% \section{MQCNN Architecture}
% \label{sxn:mqcnn_arch}

% See Figure~\ref{fig:mqcnn_arch} for an illustration of the MQCNN architecture.

% \begin{figure}[h]
%     \centering
%     \includegraphics[scale=.40]{img/MQCNN.pdf}
%     \caption{Illustration of the MQCNN architecture.}
%     \label{fig:mqcnn_arch}
% \end{figure}

%\newpage
\section{Model Quality Diagnostics using HTSR Theory}
\label{appendix:htsr}

In this section, we describe HTSR-based model quality diagnostics in more detail. 

To model random variables with HT properties (which, for us, will be the ESDs of the weight matrices of well-trained NN models), consider the general distribution,
\begin{equation}\label{eq:pw}
  \rho(\lambda;\theta)\propto C(\lambda;\theta)\lambda^{-\alpha}\mathds{1}_{\{\lambda\ge\lambda_{\min}\}}
\end{equation}
where $\theta$ is used to identify an arbitrary collection of parameters, and $\mathds{1}_{\{E\}}$ is the indicator function of the event $E$. When $C(\lambda;\theta) \propto K(\theta)$ is constant with respect to $\lambda$, \eqref{eq:pw} is called a Power Law (PL) distribution. When $C(\lambda;\theta) \propto K(\theta)e^{-\beta\lambda}$, \eqref{eq:pw} is call an exponentially Truncated Power Law (TPL) distribution. 

For an architecture with $L$ layers, let $\mathbf{W}_l$ be the real weight matrix for layer $l$. In this paper, we are particularly interested in the following two metrics:
\begin{itemize}
  \item {\bf The $\alpha$ metric.} The $\alpha$ metric from HTSR theory is a measure of the {\em shape} of the ESDs. It is the average of the fitted PL parameters $\alpha_l$ from \eqref{eq:pw} for the eigenvalues of the matrix $\mathbf{X}_l = \mathbf{W}_l^T\mathbf{W}_l$. Each $\alpha_l$ is obtained by minimizing the KS distance between the ESD of $\mathbf{X}_l$ and the PL density~\eqref{eq:pw}. Each $\alpha_l$ can be interpreted as the shape of the spectrum of the corresponding layer. In our evaluations, since some layer-level ESDs are not even close to (T)PL (and hence the parameter $\alpha_l$ can be meaningless), we only average over layers whose ESDs appear to be (T)PL.
  \item {\bf Stable Rank.} The stable rank is a norm-adjusted measure of the {\em scale} of the ESDs:
  \[
    \mbox{stable rank} = \frac{1}{L}\sum_{l=1}^L \frac{\|\mathbf{W}_l\|_F^2}{\|\mathbf{W}_l\|_2^2}.
  \]
  It is strongly predictive of model quality in the NLP domain~\cite{yang2022evaluating}.
\end{itemize}

In the rest of this section, we provide detailed diagnostics using the HT shape and scale of layer-level ESDs. We split our evaluations into three parts, and we show that HTSR metrics
\begin{itemize}
  \item are indicative of which one between GPT-2 and Flan-T5-small is better suited for our forecasting task (see Figure~\ref{fig:gpt2_vs_flant5_esd} and Figure~\ref{fig:gpt2_vs_flant5_test_loss_and_htsr_metric});
  \item identify layer-level anomalies of linear embedding/decoding layers and suggest MLP embedding/decoding (see Figure~\ref{fig:linear_vs_mlp_esd} and Figure~\ref{fig:linear_vs_mlp_test_loss_and_htsr_metric}); and
  \item verify the advantage of FPT over non-FPT baseline (see Figure~\ref{fig:fpt_vs_no_fpt_esd} and Figure~\ref{fig:fpt_vs_no_fpt_test_loss_and_htsr_metric}).
\end{itemize}

%\vspace{-1mm}
\subsection{GPT-2 vs Flan-T5}\label{sec:esd_gpt2_vs_flant5}
%\vspace{-1mm}

\begin{figure}[ht!]
  \centering
  \begin{subfigure}{0.33\textwidth}
    \centering
    \includegraphics[width=\textwidth]{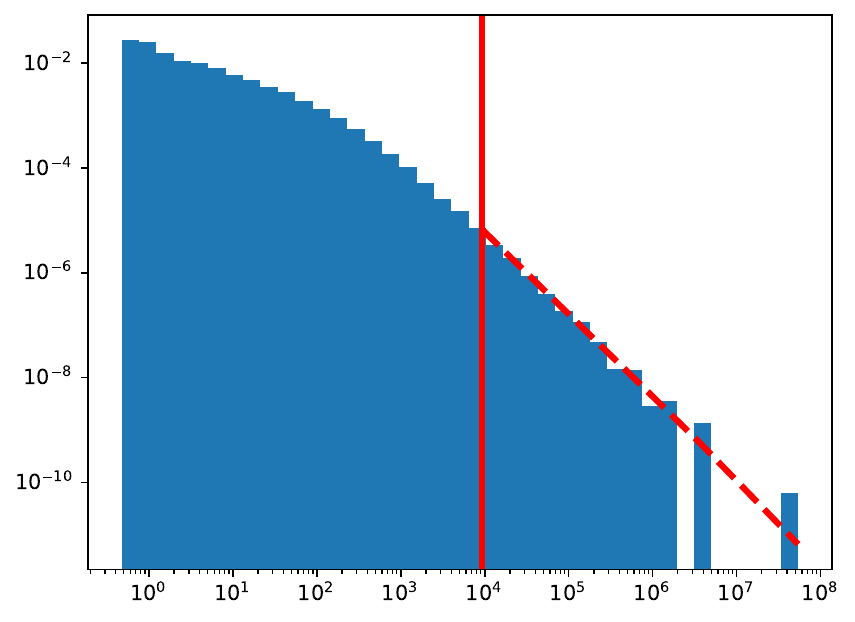}
  \end{subfigure}%
  \begin{subfigure}{0.33\textwidth}
    \centering
    \includegraphics[width=\textwidth]{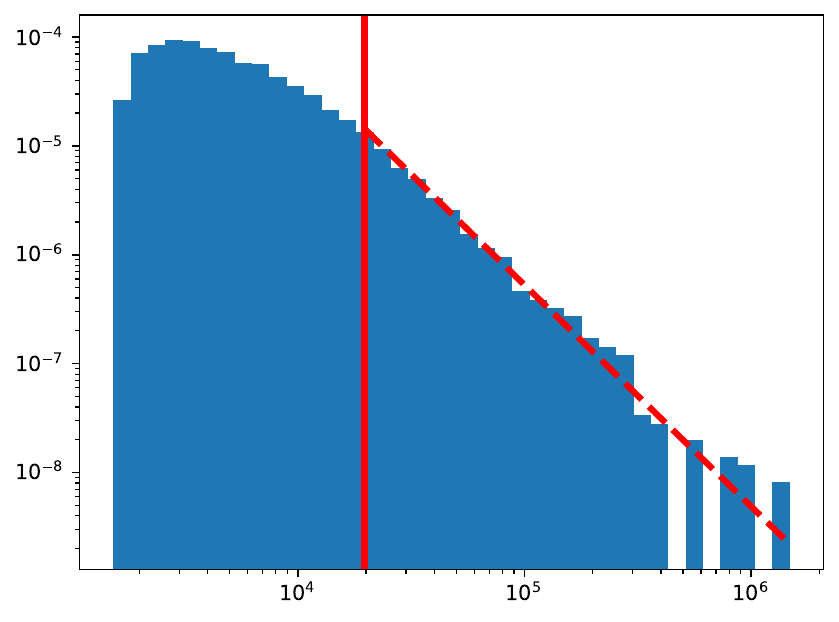}
  \end{subfigure}%
  \begin{subfigure}{0.33\textwidth}
    \centering
    \includegraphics[width=\textwidth]{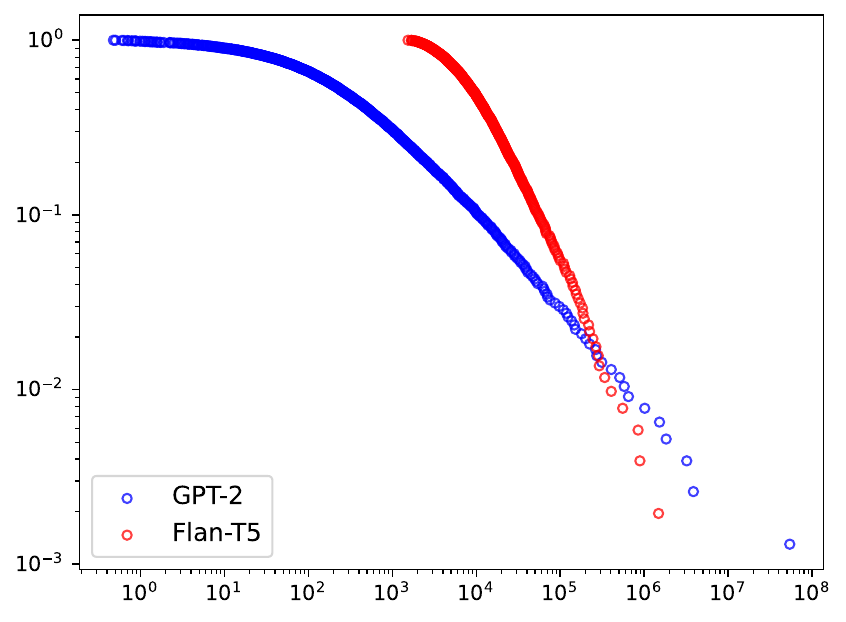}
  \end{subfigure}

  \begin{subfigure}{0.33\textwidth}
    \centering
    \includegraphics[width=\textwidth]{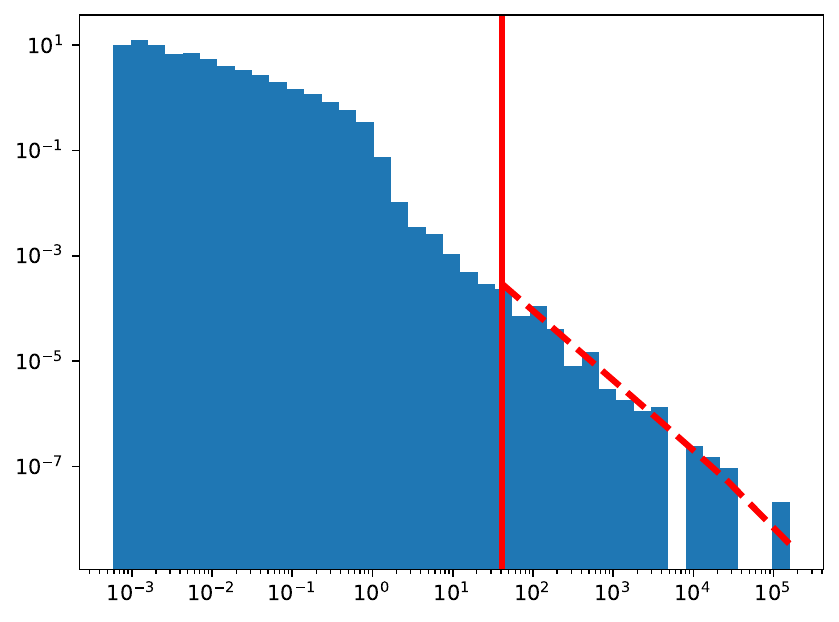}
  \end{subfigure}%
  \begin{subfigure}{0.33\textwidth}
    \centering
    \includegraphics[width=\textwidth]{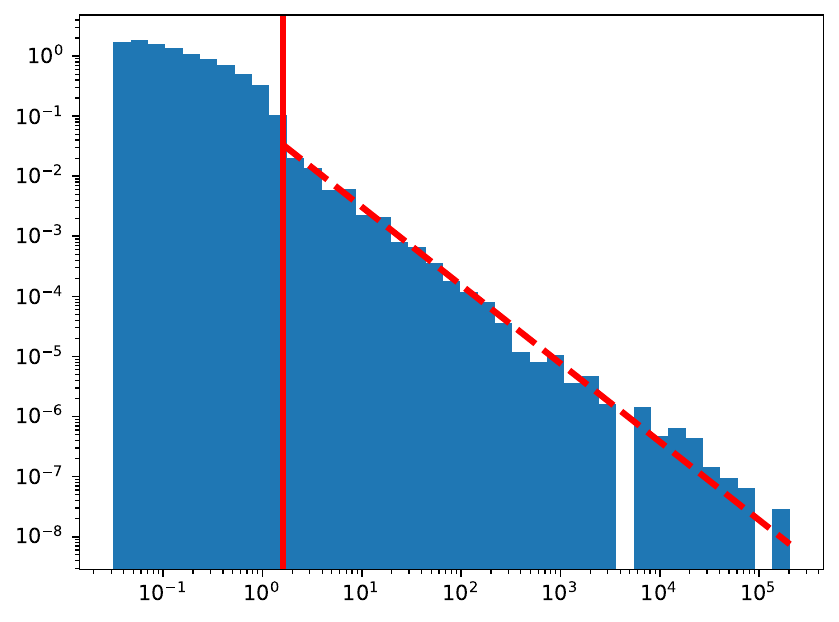}
  \end{subfigure}%
  \begin{subfigure}{0.33\textwidth}
    \centering
    \includegraphics[width=\textwidth]{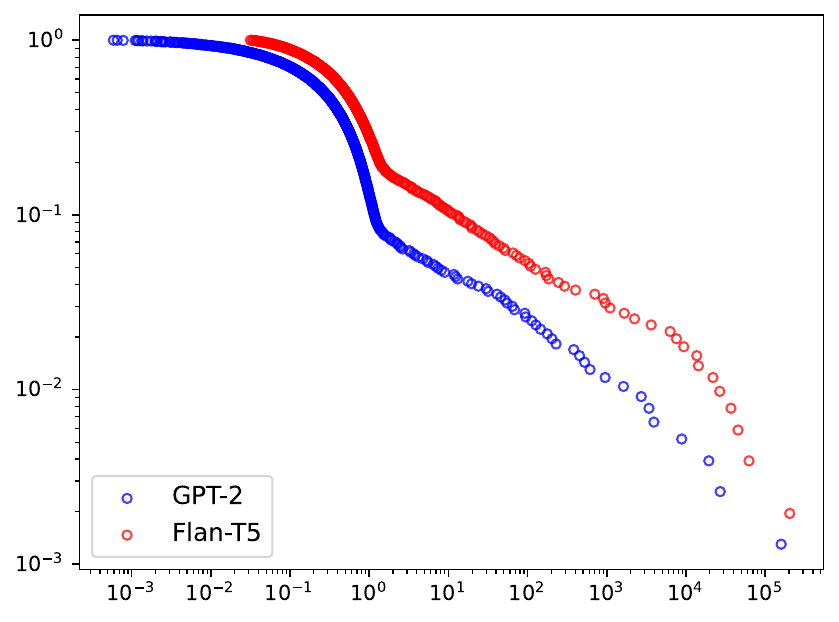}
  \end{subfigure}
  \caption{ESD of the embedding layer (top) and and the output layer (bottom), when the base FPT is GPT-2 (left) and Flan-T5 (middle). The input and output layers are liner. Red vertical lines correspond to the $\lambda_{\min}$ parameter of the PL distribution in \eqref{eq:pw} chosen by minimizing the KS distance, and red dashed lines represent the PL distribution for the fitted $\alpha$, as in previous work~\cite{martin2021predicting}. Additionally, the tail of the complementary cumulative distribution function (CCDF) is shown for each layer (right). For the embedding layer, the CCDFs for both FPTs tend to follow a PL tail.  However, the CCDF that corresponds to Flan-T5 has a much sleeper slope (i.e., better $\alpha$ metric) than the CCDF that corresponds to GPT-2. We investigate how HTSR metrics (e.g., $\alpha$) are predictive of model quality and illustrate details in Figure~\ref{fig:gpt2_vs_flant5_test_loss_and_htsr_metric}. For both architectures, the CCDFs for the output layer have a convex kink around $10^4$, and overall, they do not demonstrate a strong evidence of (T)PL tails. In Figure~\ref{fig:linear_vs_mlp_esd}, we show that replacing the linear output layer with an MLP removes the kink and results in an ESD that demonstrates (T)PL tail.}
  \label{fig:gpt2_vs_flant5_esd}
\end{figure}

\begin{figure}[ht!]
  \centering
  \begin{subfigure}{0.33\textwidth}
    \centering
    \includegraphics[width=\textwidth]{plots/tsfpt_flant5_linear_500k_100epoch_embedding_0_future_hist.pdf}
  \end{subfigure}%
  \begin{subfigure}{0.33\textwidth}
    \centering
    \includegraphics[width=\textwidth]{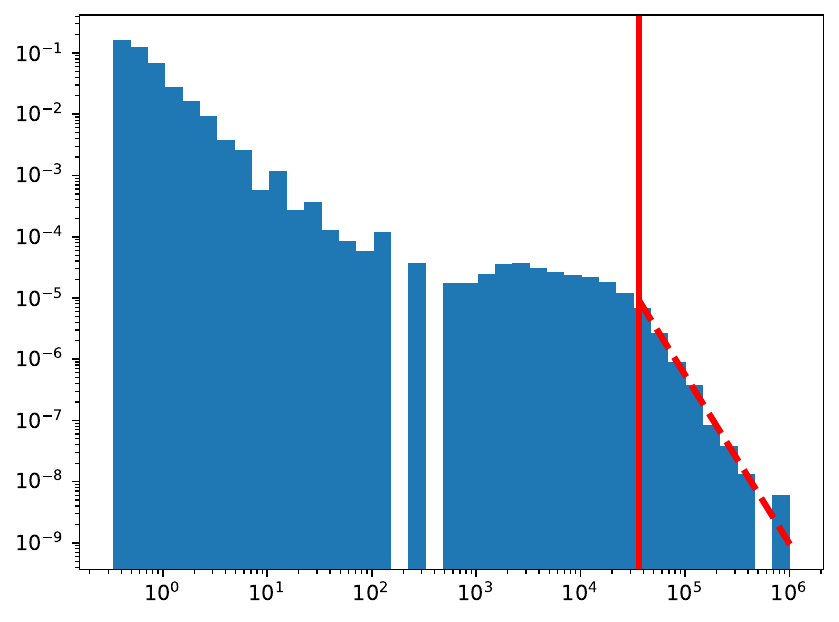}
  \end{subfigure}%
  \begin{subfigure}{0.33\textwidth}
    \centering
    \includegraphics[width=\textwidth]{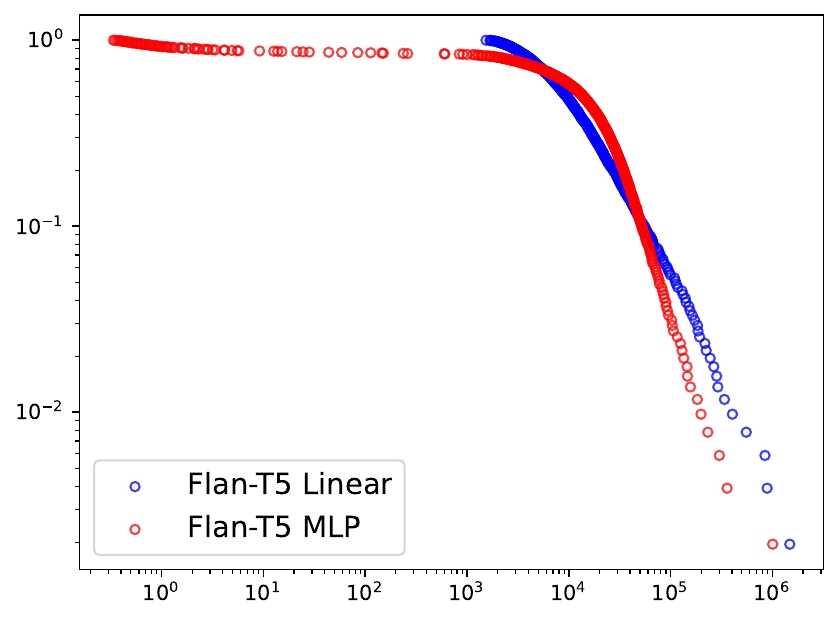}
  \end{subfigure}

  \begin{subfigure}{0.33\textwidth}
    \centering
    \hspace{\textwidth}
  \end{subfigure}%
  \begin{subfigure}{0.33\textwidth}
    \centering
    \includegraphics[width=\textwidth]{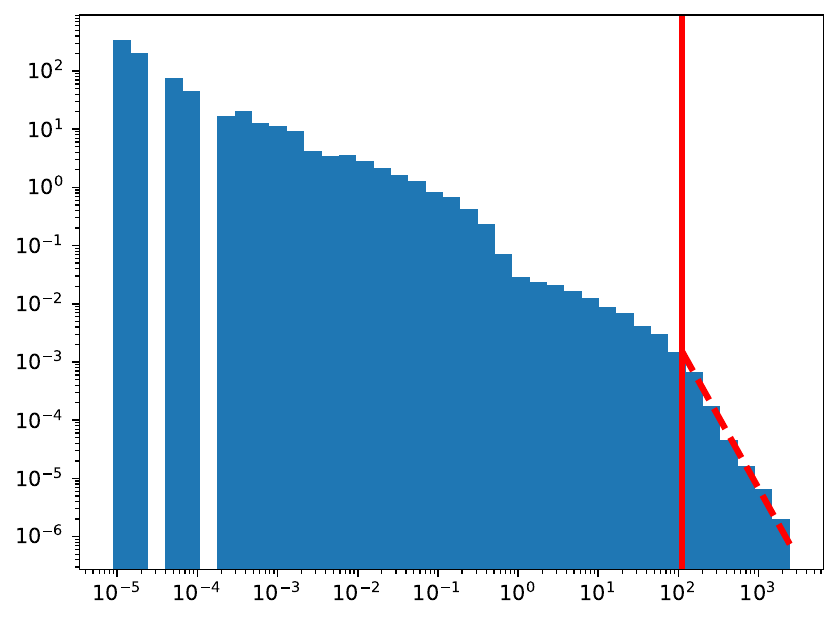}
  \end{subfigure}%
  \begin{subfigure}{0.33\textwidth}
    \centering
    \includegraphics[width=\textwidth]{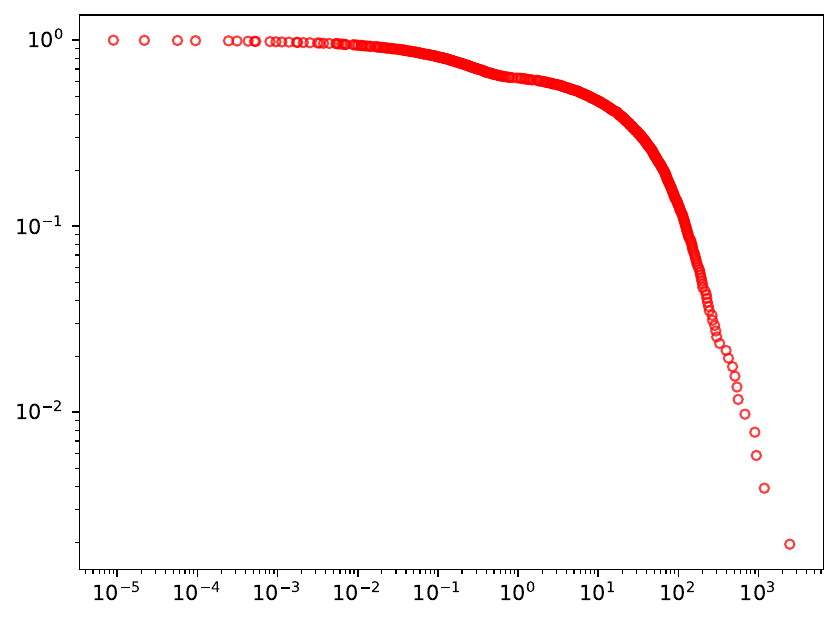}
  \end{subfigure}

  \begin{subfigure}{0.33\textwidth}
    \centering
    \includegraphics[width=\textwidth]{plots/tsfpt_flant5_linear_500k_100epoch_output_linear_future_hist.pdf}
  \end{subfigure}%
  \begin{subfigure}{0.33\textwidth}
    \centering
    \includegraphics[width=\textwidth]{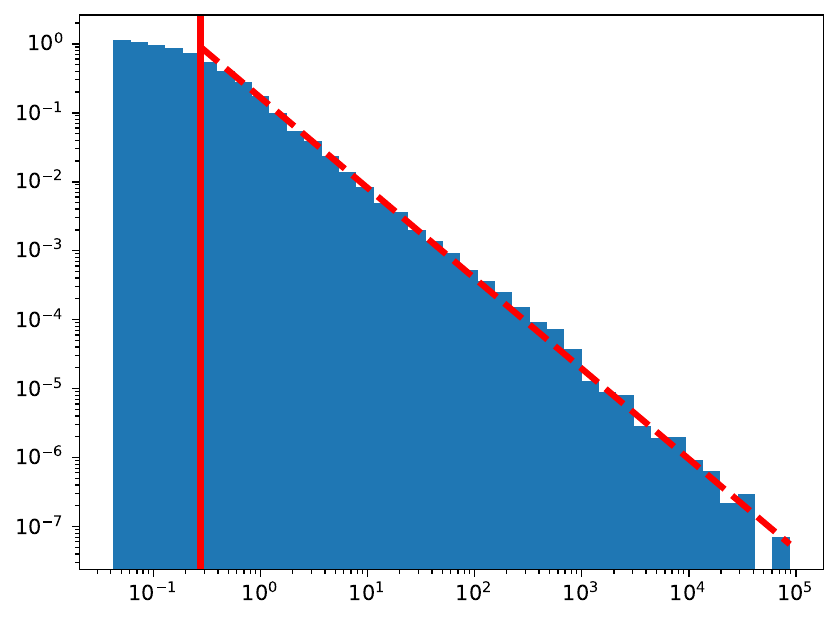}
  \end{subfigure}%
  \begin{subfigure}{0.33\textwidth}
    \centering
    \includegraphics[width=\textwidth]{plots/tsfpt_flant5_500k_100epoch_mlp_vs_linear_output_linear_future.pdf}
  \end{subfigure}
%  \begin{subfigure}{0.45\textwidth}
%    \centering
%    \includegraphics[width=\textwidth]{plots/tsfpt_flant5_500k_100epoch_mlp_vs_linear_output_linear_decoder_only.pdf}
%    \caption{Output layer (dim: 832 x 512)}
%  \end{subfigure}%
%  ~
%  \begin{subfigure}{0.45\textwidth}
%    \centering
%    \includegraphics[width=\textwidth]{plots/tsfpt_gpt2_500k_100epoch_mlp_vs_linear_output_linear_future.pdf}
%    \caption{Output layer (dim: 832 x 768)}
%  \end{subfigure}%
  \caption{ESD of the first embedding layer (top), the second embedding layer (middle, this layer is specific for MLP, and hence there is none on the left), and the output layer (bottom), when the embedding/decoding layer is linear (left) and 2-layer MLP (middle). Red vertical lines correspond to the $\lambda_{\min}$ parameter of the PL distribution in \eqref{eq:pw} chosen by minimizing the KS distance, and red dashed lines represent the PL distribution for the fitted $\alpha$. Additionally, the tail of the complementary cumulative distribution function (CCDF) is shown for each layer (right). CCDFs of the architecture that use MLP embedding/decoding not only demonstrate (T)PL tails, but also have steeper slopes (i.e., better $\alpha$ metric) than the CCDFs of the architecture that use linear embedding/decoding. The close relationship between HTSR metrics (e.g., the $\alpha$ metric) and model quality is illustrated in Figure~\ref{fig:linear_vs_mlp_test_loss_and_htsr_metric}.}
  \label{fig:linear_vs_mlp_esd}
\end{figure}

Here, we demonstrate that HTSR metrics are predictive of model accuracy for varying FPTs. We focus on architectures that add linear input and output layers to FPTs. We analyze the layer-level ESDs when the FPT is GPT-2 and Flan-T5, repseictively. Figure~\ref{fig:gpt2_vs_flant5_esd} shows the layer-level ESDs for both architectures and for both input and output layers. We also plot the complementary cumulative distribution function (CCDF) for each layer. If the ESD is PL or approximately PL, then the corresponding tail CCDF will be linear or approximately linear. Observe that, for the output layer, the CCDFs for both architectures have a convex kink around $10^4$, and overall, they do not demonstrate clear (T)PL tails. 
In the Section \ref{sec:linear_vs_mlp}
we show that replacing the linear output layer with an MLP removes the kink and results in an ESD that demonstrates (T)PL tail (see Figure~\ref{fig:linear_vs_mlp_esd}, bottom right).
Our empirical results (in Section~\ref{sxn:main_results}) show that the resulting model is much better.

\begin{figure}[ht!]
  \centering
  \begin{subfigure}{0.4\textwidth}
    \centering
    \includegraphics[width=\textwidth]{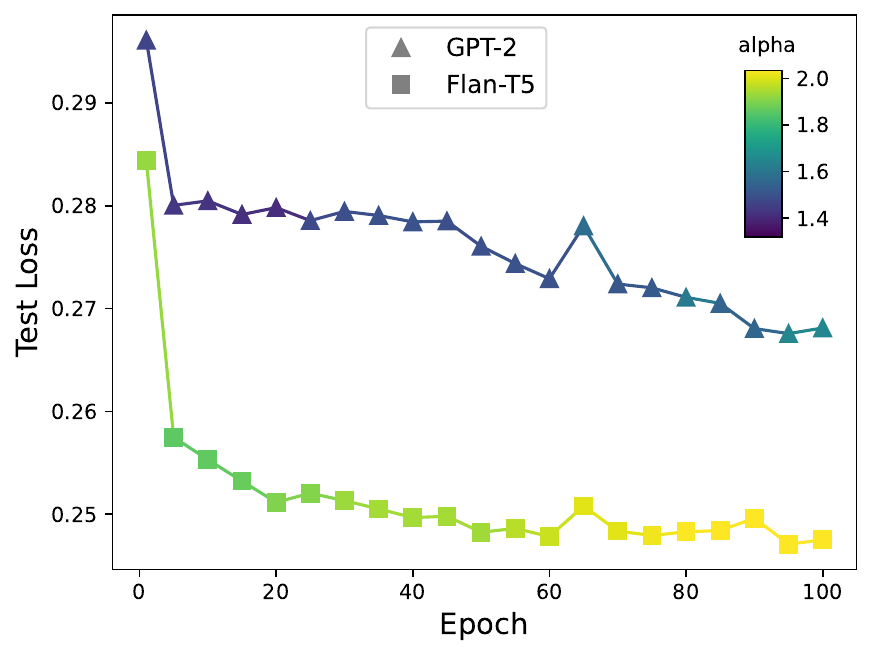}
  \end{subfigure}%
  ~
  \begin{subfigure}{0.4\textwidth}
    \centering
    \includegraphics[width=\textwidth]{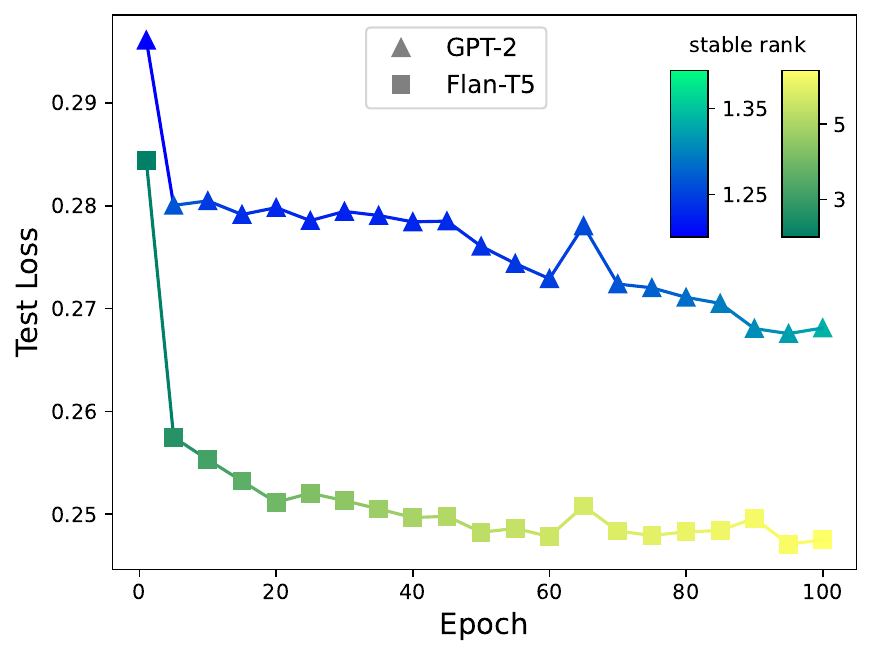}
  \end{subfigure}
  \caption{{\bf HTSR metrics predict forecasting accuracy across architectures (varying base FPTs) and within architecture across epochs.} P50 quantile test loss by epoch for GPT-2- and Flan-T5-based architectures. Markers and lines are colored according to the $\alpha$ metric (left) and the stable rank metric (right). Within the same architecture over different epochs, a higher metric value generally results in a higher accuracy; at a fixed epoch between different architectures, a higher metric value generally results in a higher accuracy.}
  \label{fig:gpt2_vs_flant5_test_loss_and_htsr_metric}
\end{figure}

For the embedding layer, the CCDF for both FPTs tend to follow a PL tail.
However, the CCDF that corresponds to Flan-T5 has a much sleeper slope (i.e., better $\alpha$ metric) than the CCDF that corresponds to GPT-2. To investigate how HTSR metrics are predictive of model quality and forecasting accuracy, in Figure~\ref{fig:gpt2_vs_flant5_test_loss_and_htsr_metric} we plot test loss over 100 epochs, and we color the loss curve by the $\alpha$ metric and the stable rank, respectively. Figure~\ref{fig:gpt2_vs_flant5_test_loss_and_htsr_metric} shows that both metrics are highly correlative with forecasting accuracy: within the same architecture over different epochs, a higher metric value generally results in a higher accuracy; and at a fixed epoch between different architectures, a higher metric value generally results in a higher accuracy. Note that, in Figure~\ref{fig:gpt2_vs_flant5_test_loss_and_htsr_metric}, the left plot involves some fitted $\alpha$ values that are less than 2. They correspond to PL distribution \eqref{eq:pw} with an exponent that is less than $2$. Such PL distribution does not have a finite mean and may cause issues in statistical diagnostics. Therefore, an $\alpha$ metric value that is less than 2 may not be reliable. Nonetheless, the results in both plots in Figure~\ref{fig:gpt2_vs_flant5_test_loss_and_htsr_metric} are consistent.

%\begin{figure}[ht!]
%  \centering
%  \begin{subfigure}{0.33\textwidth}
%    \centering
%    \includegraphics[width=\textwidth]{plots/tsfpt_gpt2_500k_evolution_embedding_0_future.pdf}
%  \end{subfigure}%
%  \begin{subfigure}{0.33\textwidth}
%    \centering
%    \includegraphics[width=\textwidth]{plots/tsfpt_flant5_500k_evolution_embedding_0_future.pdf}
%  \end{subfigure}%
%  \begin{subfigure}{0.33\textwidth}
%    \centering
%    \includegraphics[width=\textwidth]{plots/tsfpt_flant5_vs_gpt2_500k_alpha_evolution_embedding_0_future.pdf}
%  \end{subfigure}
%  \begin{subfigure}{0.33\textwidth}
%    \centering
%    \includegraphics[width=\textwidth]{plots/tsfpt_gpt2_500k_evolution_output_linear_future.pdf}
%  \end{subfigure}%
%  \begin{subfigure}{0.33\textwidth}
%    \centering
%    \includegraphics[width=\textwidth]{plots/tsfpt_flant5_500k_evolution_output_linear_future.pdf}
%  \end{subfigure}%
%  \begin{subfigure}{0.33\textwidth}
%    \centering
%    \includegraphics[width=\textwidth]{plots/tsfpt_flant5_vs_gpt2_500k_alpha_evolution_output_linear_future.pdf}
%  \end{subfigure}
%  \caption{\shenghao{Let's remove this figure (since these plots can be confusing)?}Evolution of top 100 eigenvalues for GPT-2 (left) and Flan-T5 (middle) for the embedding layer (top) and the output layer (bottom). We also plot the evolution of $\alpha$ (right).}
%\end{figure}

%\clearpage

%\vspace{-1mm}
\subsection{Linear vs MLP Embedding/Decoding}\label{sec:linear_vs_mlp}
%\vspace{-1mm}

\begin{figure}[ht!]
  \centering
  \begin{subfigure}{0.4\textwidth}
    \centering
    \includegraphics[width=\textwidth]{plots/tsfpt_flant5_linear_vs_mlp_500k_test_loss_color_by_alpha_embedding_layers_future.pdf}
  \end{subfigure}%
  ~
  \begin{subfigure}{0.4\textwidth}
    \centering
    \includegraphics[width=\textwidth]{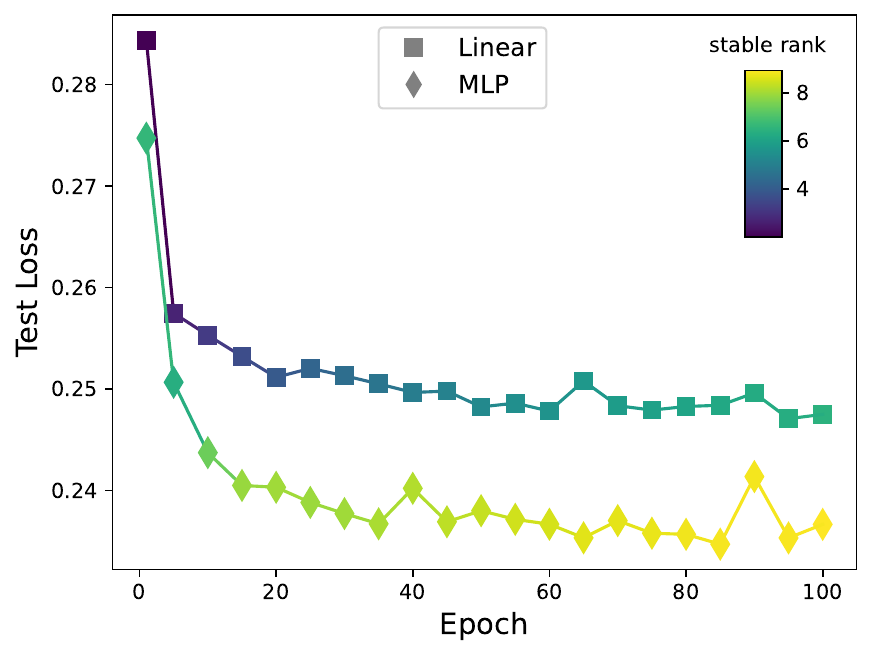}
  \end{subfigure}
  \caption{{\bf HTSR metrics predict forecasting accuracy across architectures (varying input/output layers) and within architecture across epochs.} P50 quantile test loss by epoch for architectures with a Linear and MLP embedding/decoding layer, respectively. Markers and lines are colored according to the $\alpha$ metric (left) and the stable rank metric (right).}
  \label{fig:linear_vs_mlp_test_loss_and_htsr_metric}
\end{figure}

Here, for illustration purpose, we fix the base FPT to Flan-T5-small, and we analyze the ESDs when the embedding and decoding layers are linear and MLP, respectively. The bottom right plot in Figure~\ref{fig:linear_vs_mlp_test_loss_and_htsr_metric} shows that an MLP decoding layer fixes the anomaly (i.e., the convex kink around $10^4$) in the ESD when the decoding layer is linear. The resulting ESD with MLP decoding layer exhibit a clear (T)PL tail. Not surprisingly, Flan-T5 with MLP input and output layers results in much better forecasting accuracy, as demonstrated both in Table~\ref{tab:500k} and in Figure~\ref{fig:linear_vs_mlp_test_loss_and_htsr_metric}. Again, Figure~\ref{fig:linear_vs_mlp_test_loss_and_htsr_metric} shows that both the $\alpha$ metric and the stable rank are strongly predictive of forecasting
accuracy: within the same architecture over different epochs, a higher metric value generally results in a higher accuracy; at a fixed epoch between different architectures, a higher metric value generally results in a higher accuracy.

%\vspace{-1mm}
\subsection{FPT vs Non-FPT Baseline}\label{sec:esd_fpt_vs_non_fpt}
%\vspace{-1mm}

\begin{figure}[ht!]
  \centering
  \begin{subfigure}{0.33\textwidth}
    \centering
    \includegraphics[width=\textwidth]{plots/tsfpt_flant5_linear_500k_100epoch_embedding_0_future_hist.pdf}
  \end{subfigure}%
  \begin{subfigure}{0.33\textwidth}
    \centering
    \includegraphics[width=\textwidth]{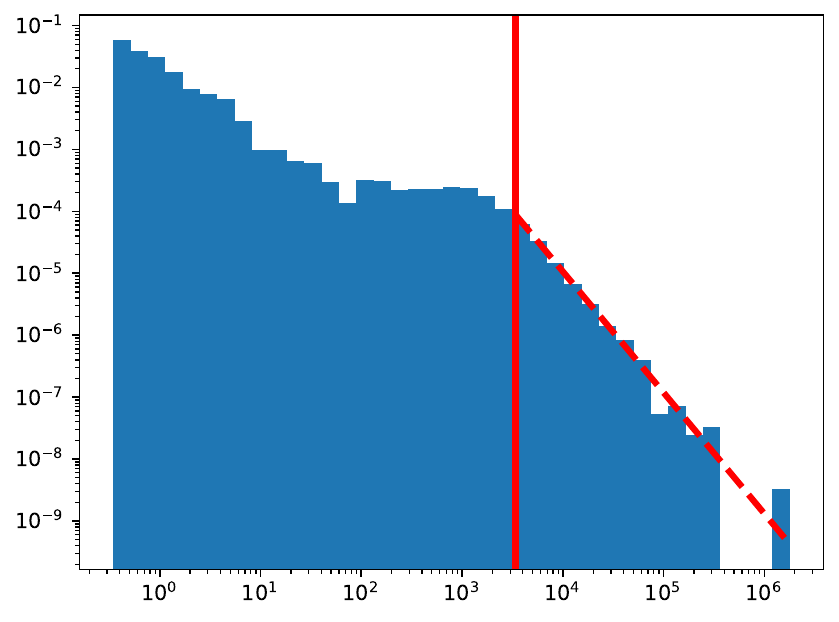}
  \end{subfigure}%
  \begin{subfigure}{0.33\textwidth}
    \centering
    \includegraphics[width=\textwidth]{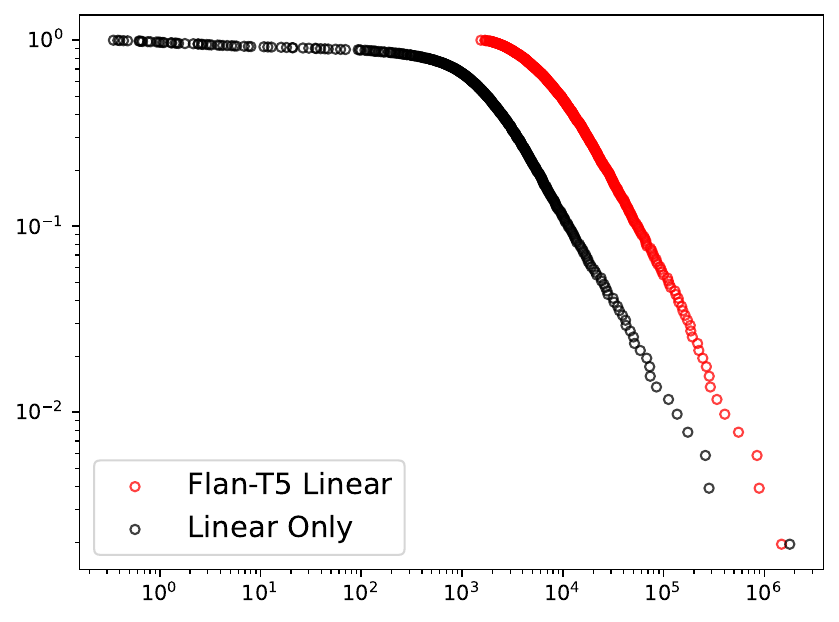}
  \end{subfigure}

  \begin{subfigure}{0.33\textwidth}
    \centering
    \includegraphics[width=\textwidth]{plots/tsfpt_flant5_linear_500k_100epoch_output_linear_future_hist.pdf}
  \end{subfigure}%
  \begin{subfigure}{0.33\textwidth}
    \centering
    \includegraphics[width=\textwidth]{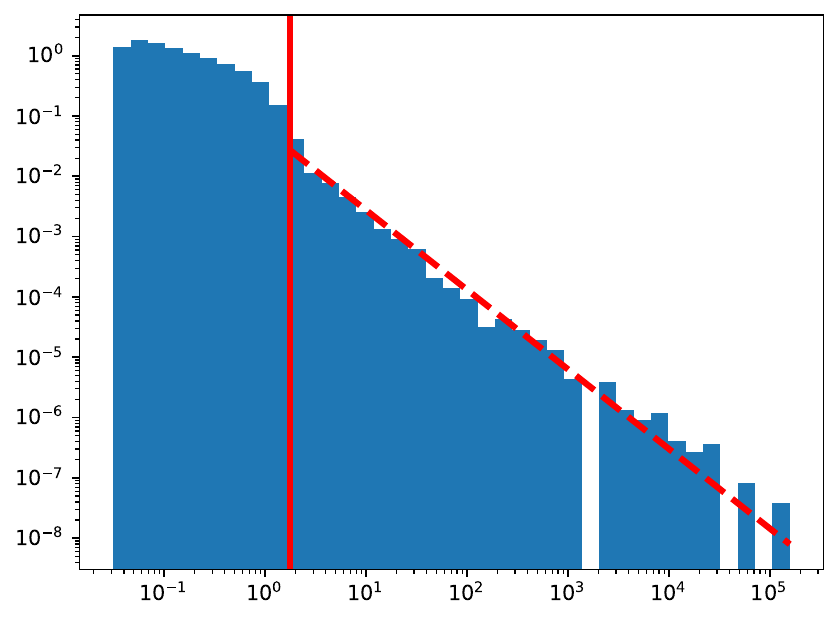}
  \end{subfigure}
  \begin{subfigure}{0.33\textwidth}
    \centering
    \includegraphics[width=\textwidth]{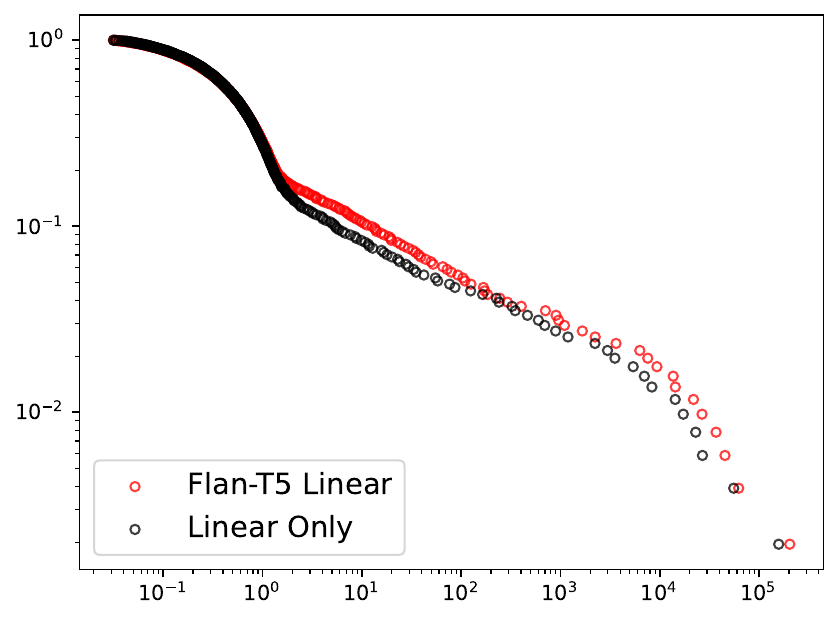}
  \end{subfigure}
  \caption{ESD of the embedding layer (top) and the output layer (bottom), when FPT is included (left) and not included (middle) in the architecture. For this figure, Flan-T5-small is used as the FPT, and we analyze architectures that adopt linear input and output layers. Red vertical lines correspond to the $\lambda_{\min}$ parameter of the PL distribution in \eqref{eq:pw} chosen by minimizing the KS distance, and red dashed lines represent the PL distribution for the fitted $\alpha$. Additionally, the tail of the complementary cumulative distribution function (CCDF) is shown for each layer (right).
  }
  \label{fig:fpt_vs_no_fpt_esd}
\end{figure}

\begin{figure}[ht!]
  \centering
  \begin{subfigure}{0.4\textwidth}
    \centering
    \includegraphics[width=\textwidth]{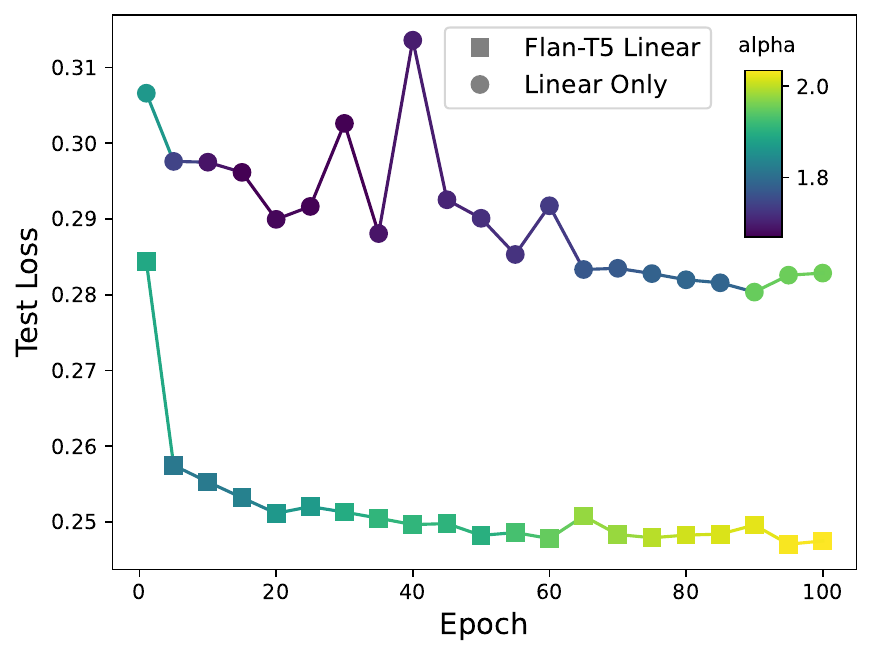}
  \end{subfigure}%
  ~
  \begin{subfigure}{0.4\textwidth}
    \centering
    \includegraphics[width=\textwidth]{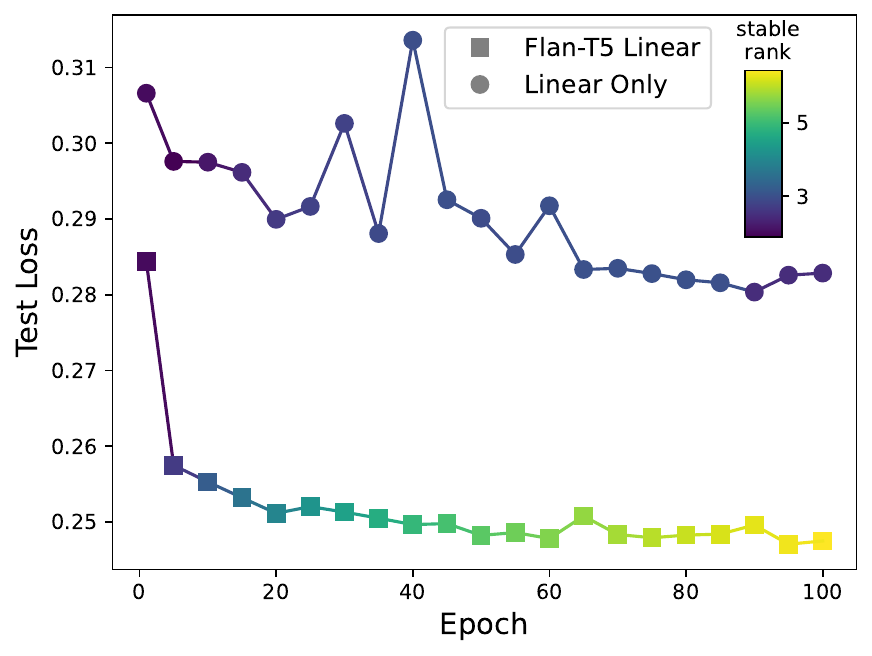}
  \end{subfigure}%
  \caption{{\bf HTSR metrics predict forecasting accuracy across architectures (FPT vs non-FPT) and within architecture across epochs.} P50 quantile test loss by epoch for architectures with a Linear and MLP embedding/decoding layer, respectively. Markers and lines are colored according to the $\alpha$ metric (left) and the stable rank metric (right). }
  \label{fig:fpt_vs_no_fpt_test_loss_and_htsr_metric}
\end{figure}

Here, we show the ESD, PL fit, and CCDF of the embedding layer and output layers of the learned maps to and from Flan-T5 relative to our ``Linear Only'' baseline.  
See Figure~\ref{fig:fpt_vs_no_fpt_esd}.
While the CCDFs for the output layer are similar, the ESD of the embedding layer of the architecture that uses Flan-T5 has a better $\alpha$ metric than that of the baseline architecture, which is just a linear network. We further demonstrate how HTSR metrics are predictive of model quality in this setting.
See Figure~\ref{fig:fpt_vs_no_fpt_test_loss_and_htsr_metric}.
As we saw before, we see that within the same architecture over different epochs, higher HTSR metric values generally correspond to higher accuracy; at a fixed epoch between different architectures, a higher metric value generally corresponds to higher accuracy.

%\newpage
\section{Illustration of Multivariate Patching}

\michael{Is this subsection referred to anywhere in the text?  We should refer to it if we keep it.  Also, if we are going to keep this section and figure, then I think we need a better explanation of that the figure is doing, since I'm not sure really what it is showing.  I like the idea of including it, if only for eye candy, since one of our selling points is multivariate.  But we can remove if we don't have a better way to explain it.}

\begin{figure}[h]
    \centering
    \hspace{-.3in}
    \includegraphics[scale=0.35]{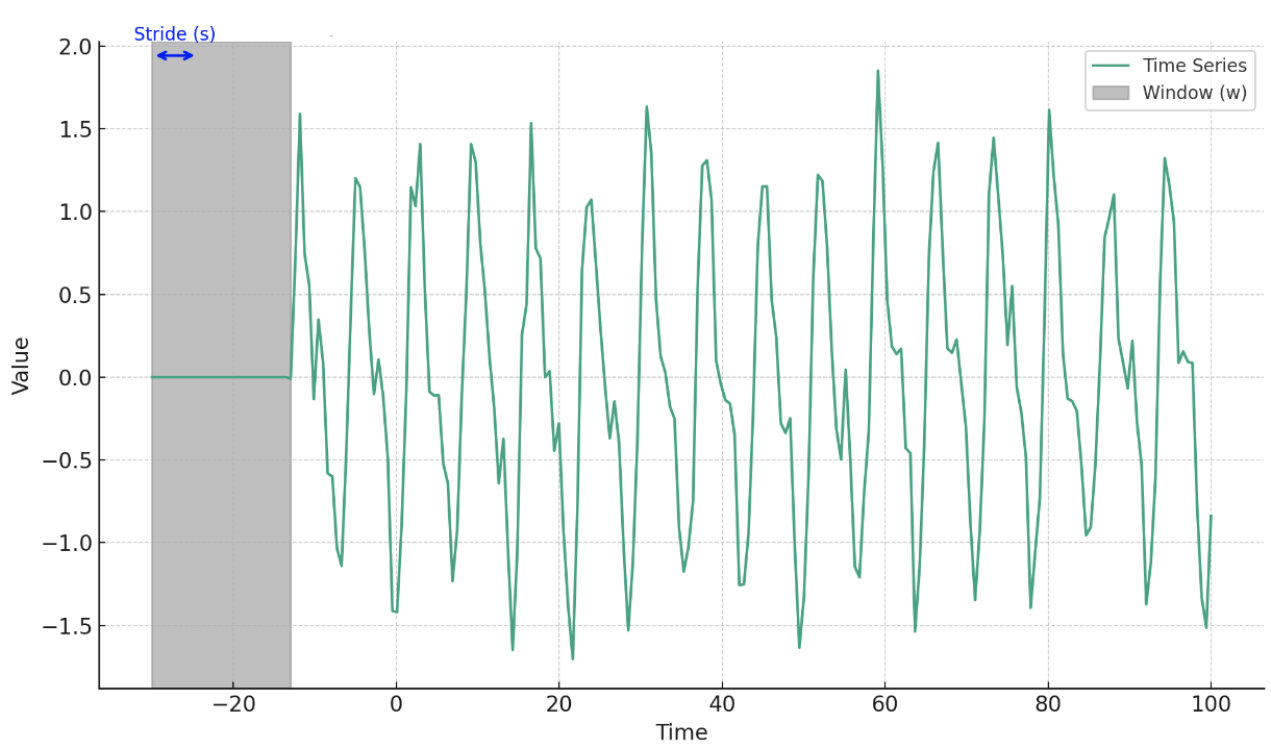}
    \hspace{.3in}
    \includegraphics[scale=0.35]{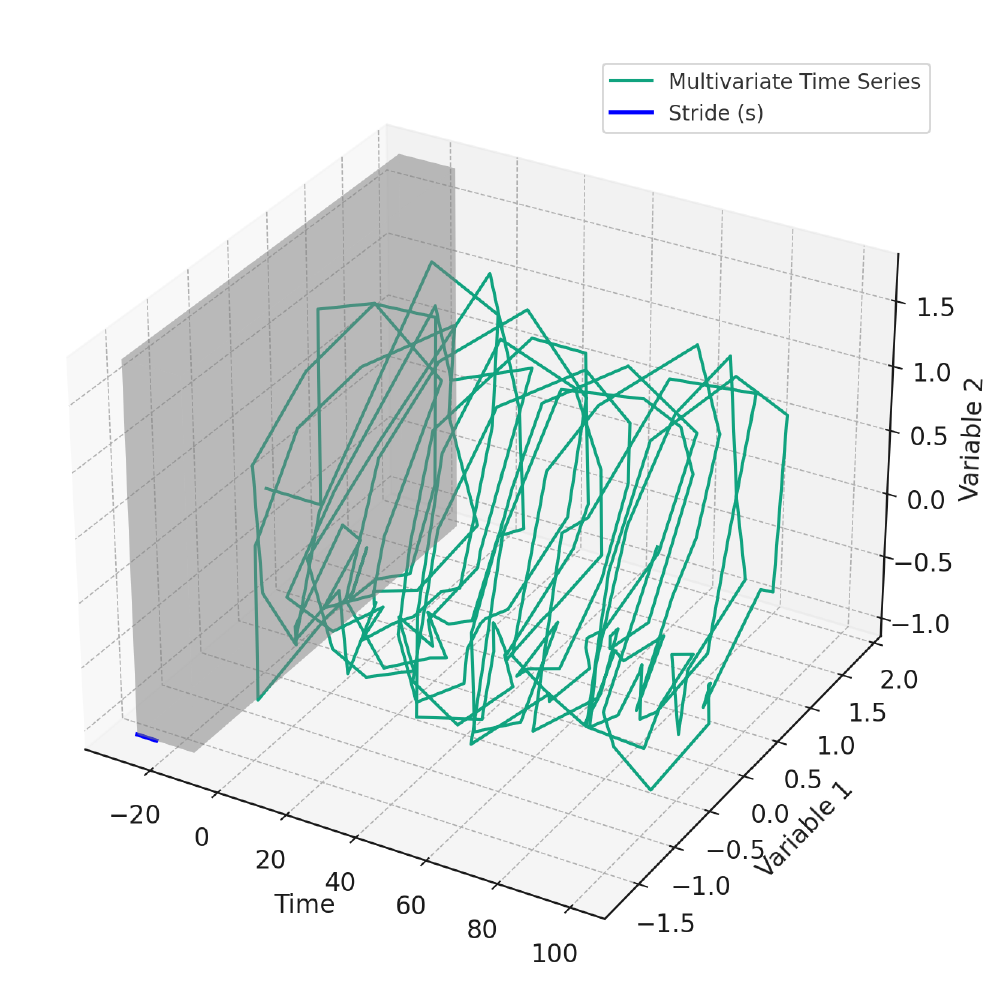}
    \caption{Representation of univariate patching \cite{nie2023time} for time series (left) and multivariate patching for time series (right). In the multivariate setting, the time series is first patched, and then flattened across the patch and covariates prior to embedding.}
    \label{fig:multi-patch}
\end{figure}

In this section, we show a visual representation of univariate and multivariate patching for time series. 
See Figure~\ref{fig:multi-patch}. 
In the univariate setting, the time series are first left-padded with zeros and then ``patched,'' i.e. flattened according to window size, to ensure causality. The same thing is done in the multivariate setting.

\end{document}